\documentclass{article} 
\usepackage{iclr2027_conference,times}

\usepackage{amsmath,amsfonts,bm}

\def\eqref#1{equation~\ref{#1}}

\def\1{\bm{1}}

\DeclareMathAlphabet{\mathsfit}{\encodingdefault}{\sfdefault}{m}{sl}
\SetMathAlphabet{\mathsfit}{bold}{\encodingdefault}{\sfdefault}{bx}{n}

\usepackage[T1]{fontenc}
\usepackage{hyperref}
\usepackage{booktabs}
\usepackage{url}
\usepackage{tikz}
\usetikzlibrary{arrows.meta,positioning,calc,decorations.pathreplacing}
\usepackage{bm}        
\usepackage{amsmath}   
\usepackage{pgfplots}
\pgfplotsset{compat=1.18}
\usepgfplotslibrary{groupplots}
\definecolor{objA}{RGB}{235,110,40}
\definecolor{objB}{RGB}{45,120,225}
\definecolor{objC}{RGB}{40,175,85}
\title{Mask-supervised Object-centric Representation Learning with LeJEPA}
\iclrfinalcopy
\author{Jakob Geusen \& Ender Konukoglu \\
Biomedical Image Computing Group\\
ETH Zürich\\
\texttt{\{jgeusen,kender\}@vision.ee.ethz.ch} \\
}
\begin{document}
\maketitle
\lhead{Preprint} 
\begin{abstract}
Self-supervised image encoders deliver strong features for downstream tasks but need many images for training. 
A natural remedy to counter this is to make each image count for more. 
A scene contains many objects, and given masks from human annotators or an off-the-shelf segmentation model, pre-training can focus on aligning per-object rather than image-wide representations, extracting more signal from every image. 
Existing mask-supervised methods do this through reconstruction or contrastive losses that leverage negative objects. 
We instead use two separate projection spaces for the alignment. 
In a \emph{semantic space}, per-object representations from different views are aligned. To avoid collapse, instead of using negative objects, which requires category definitions, we extend the negative-free LeJEPA objective and show that its distributional anti-collapse regularizer ports naturally from whole images to the variable-sized set of objects in a scene.
In an \emph{instance space}, a contrastive loss separates per-object representations from their context and co-occurring instances, including those of the same category.
To separate object representations from their context, we copy objects and paste them into other contexts, where each pasted copy serves as an additional view of the original object.
Trained on COCO with ground-truth masks, our method outperforms image-level and mask-guided baselines on tracking (DAVIS), classification (ImageNet-1k) and re-identification (NAVI), matches the best of them on semantic segmentation (ADE20k) and keeps its lead over image-level LeJEPA and a supervision-matched alternative on COCO fractions down to 256 images.
\end{abstract}
\section{Introduction}
\begin{figure*}[h]
  \centering
  \resizebox{\textwidth}{!}{%
  \begin{tikzpicture}[
      >={Latex[length=2.2mm]},
      flow/.style={->,line width=0.7pt},
      proc/.style={draw,rounded corners,align=center,fill=black!4,line width=0.7pt},
      zbox/.style={draw,rounded corners,minimum width=1.1cm,minimum height=0.62cm,line width=0.8pt},
      lbl/.style={font=\small,align=center},
    ]
    \def\imgw{4}                             
    \pgfmathsetmacro\imgh{\imgw*2/3}
    \def\ncols{24}
    \def\nrows{16}
    \def\sep{0.02}                           
    \pgfmathsetmacro\pw{\imgw/\ncols}        
    \pgfmathsetmacro\gridw{\ncols*\pw+(\ncols-1)*\sep}
    \pgfmathsetmacro\gridh{\nrows*\pw+(\nrows-1)*\sep}
    \node[inner sep=0] (input) at (0,0)
      {\includegraphics[width=\imgw cm]{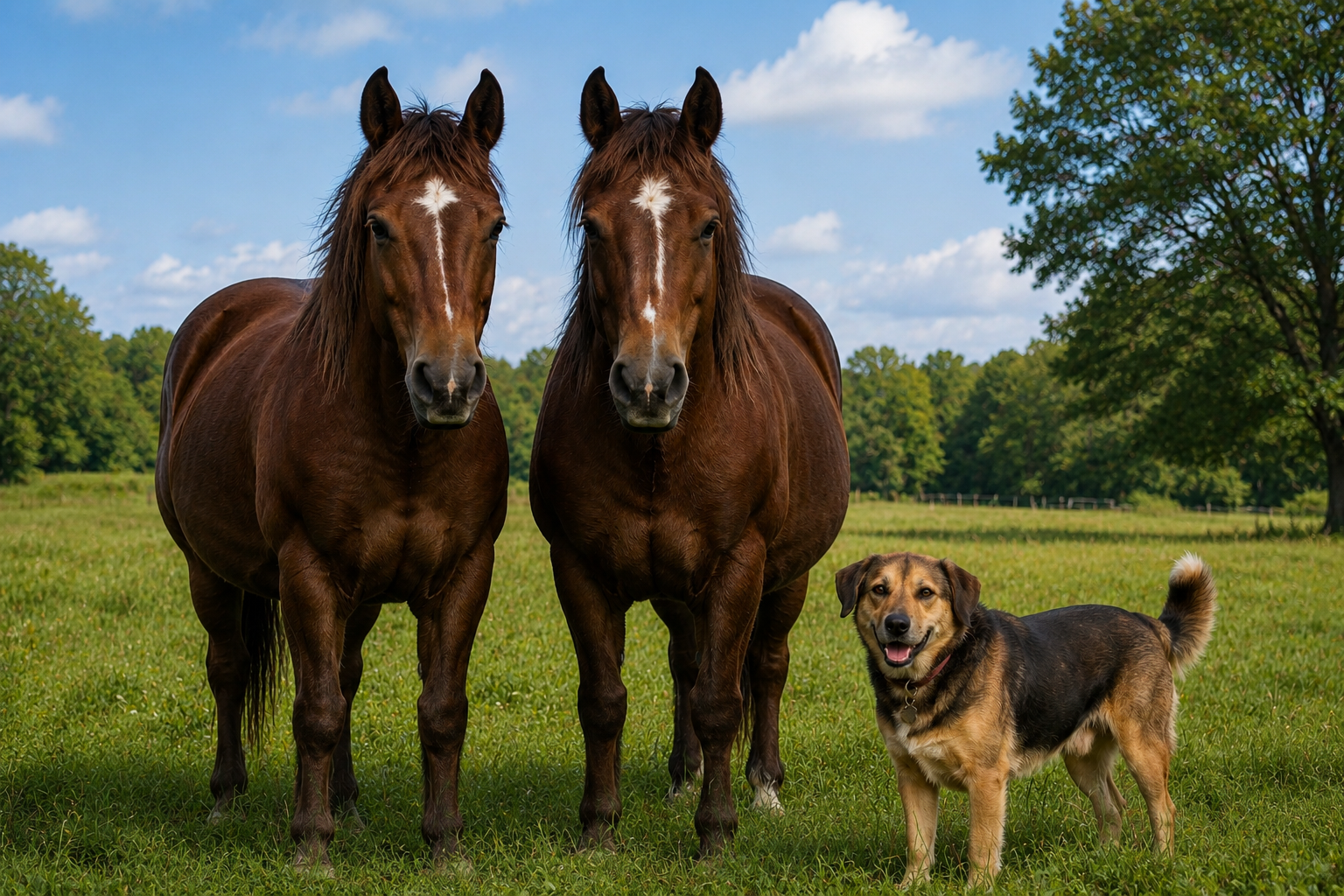}};
    \node[lbl,below=2pt of input] {input image $\mathbf{x}$};
    \def\PCx{6.0}
    \node[inner sep=0] (patched) at (\PCx,0)
      {\includegraphics[width=\gridw cm]{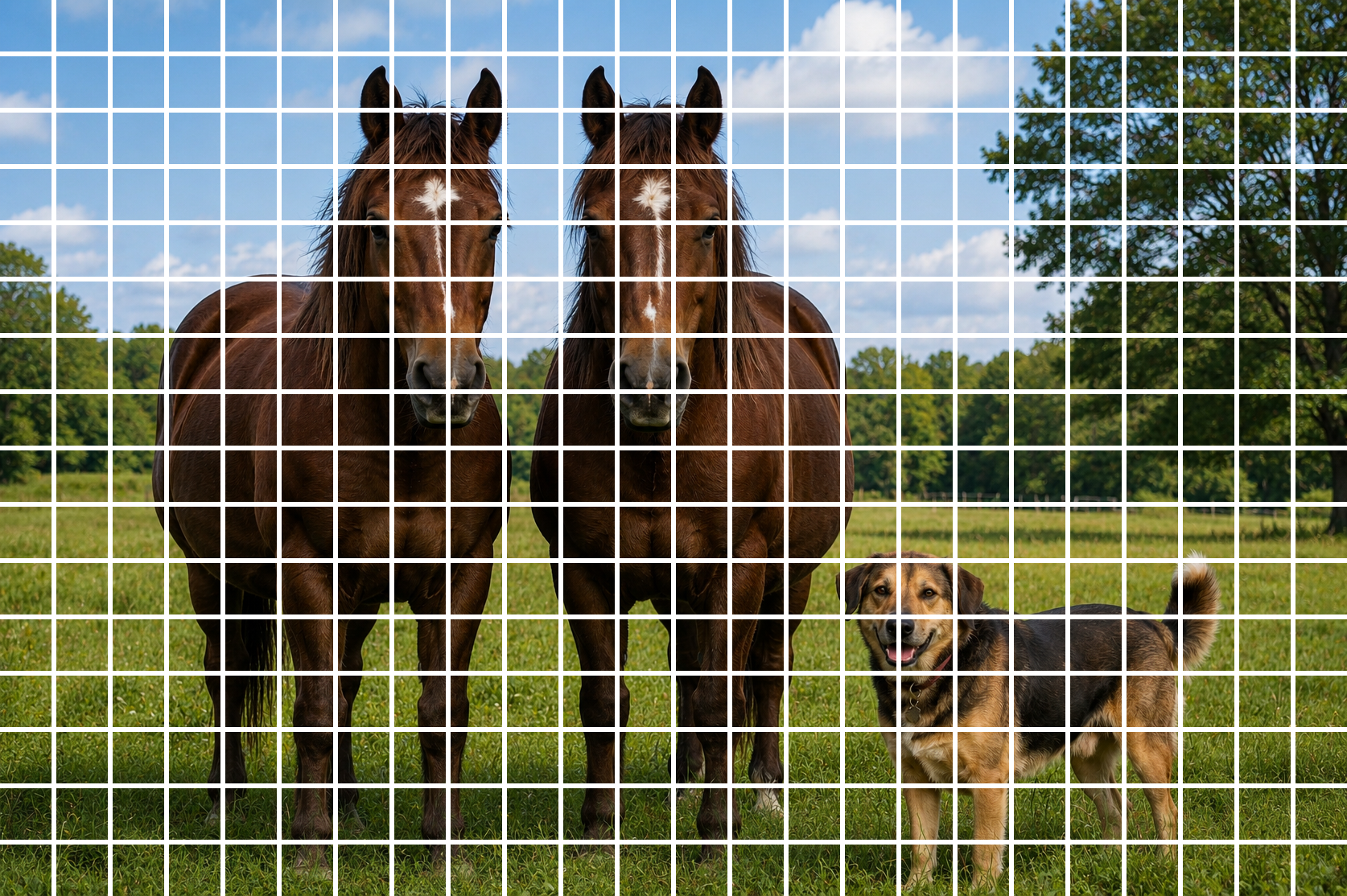}};
    \node[lbl,below=2pt of patched] {patch features $g(\mathbf{x})$};
    \def\PARTx{11.6}
    \node[inner sep=0] (part) at (\PARTx,0)
      {\includegraphics[width=\gridw cm]{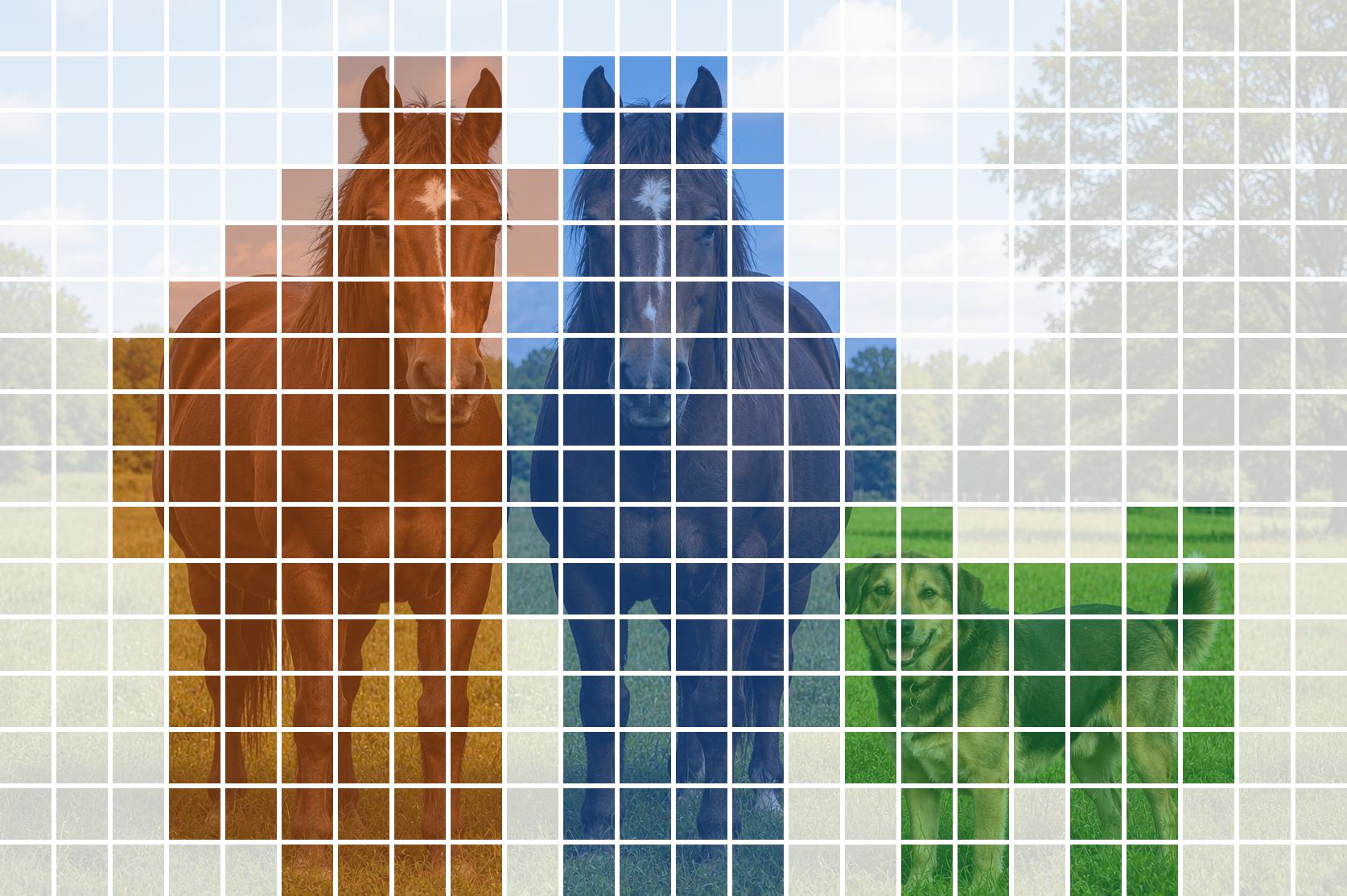}};
    \node[lbl,below=2pt of part] {masked patch features $g(\mathbf{x})$};
    \node[proc,text width=3.4cm] (mg) at (\PARTx,3.6)
      {External Mask Source};
    \draw[flow] (input.east) -- node[above,font=\small]{$g$} (patched.west);
    \draw[flow] (patched.east) -- (part.west);
    \draw[flow] (mg.south) -- node[right,font=\small]{masks} (part.north);
    \def\gw{2.8}                             
    \def\GX{17}\def\FX{20}\def\ZX{22}\def\BX{14.3}
    \foreach \i/\yy/\col in {0/2.4/objA, 1/0/objB, 2/-2.4/objC} {
      \node[draw=\col,line width=1pt,inner sep=1pt] (g\i) at (\GX,\yy)
        {\includegraphics[width=\gw cm]{resources/group_\i.png}};
      \node[proc,circle,minimum size=0.9cm] (f\i) at (\FX,\yy) {$f$};
      \node[zbox,fill=\col!22,draw=\col] (z\i) at (\ZX,\yy) {$\mathbf{z}_{\the\numexpr\i+1\relax}$};
      \draw[flow] (part.east) -- (\BX,0) |- (g\i.west);
      \draw[flow] (g\i.east) -- (f\i.west);
      \draw[flow] (f\i.east) -- (z\i.west);
    }
    \node[lbl,align=center] at (\ZX,3.4) {semantic\\object\\reps.};
    \draw[decorate,decoration={brace,amplitude=6pt}] (\ZX+0.6,-3.4) -- (\ZX-0.6,-3.4);
    \node[lbl,align=center] at (\ZX,-4.0) {{\Large $\mathcal{L}_{\mathrm{ObjectLeJEPA}}$}};
    \def\IRphi{-3.2}            
    \def\IRpred{-4.6}           
    \def\ISpace{0.95}           
    \foreach \i/\fl/\dr in {%
        0/black!12/black!55, 1/objA!22/objA, 2/objA!22/objA,
        3/objB!22/objB, 4/objC!22/objC, 5/black!12/black!55} {
      \pgfmathsetmacro\px{11.3+(\i-2.5)*\ISpace}
      \node[proc,circle,minimum size=0.78cm] (phi\i) at (\px,\IRphi) {$\phi$};
      \node[draw=\dr,fill=\fl,rounded corners,minimum width=0.72cm,minimum height=0.5cm,line width=0.8pt]
        (ip\i) at (\px,\IRpred) {};
      \draw[flow] (phi\i.south) -- (ip\i.north);
    }
    \node[font=\small] at (14.5,\IRphi) {$\cdots$};
    \node[font=\small] at (14.5,\IRpred) {$\cdots$};
    \node[lbl,font=\scriptsize,align=right,anchor=east] at (8.6,\IRpred) {instance object\\predictions};
    \draw[flow, gray, dotted] (10, 0.8) -- (phi0);
    \draw[flow, gray, dotted] (10.4, 0.1) -- (phi1);
    \draw[flow, gray, dotted] (10.95, -0.1) -- (phi2);
    \draw[flow, gray, dotted] (11.7, 0.5) -- (phi3);
    \draw[flow, gray, dotted] (12.45, -0.45) -- (phi4);
    \draw[flow, gray, dotted] (13.55, -0.65) -- (phi5);
    \draw[decorate,decoration={brace,amplitude=6pt}] (14.8,-5.1) -- (8.6,-5.1);
    \node[lbl,align=center] at (12.1,-5.6) {{\Large $\mathcal{L}_{\mathrm{instance}}$}};
  \end{tikzpicture}}
  \vspace{-6mm}
  \caption{Overview of our training framework. An exemplary (AI-generated for the demonstration \citep{chatgpt2026}) input image $\mathbf{x}$ is embedded by a Vision Transformer (ViT) backbone $g$ into per-patch features. We assume masks to be given -- here two horses and a dog (orange, blue, green). For each object, the aggregator $f$ pools its patch features into a single object representation $\mathbf{z}_k$. The proposed object-centric LeJEPA loss $\mathcal{L}_{\mathrm{ObjectLeJEPA}}$ aligns semantic object representations across augmentations, while avoiding collapse, here ideally resulting in the two horses ($\mathbf{z}_1,\mathbf{z}_2$) having similar representations while keeping the dog ($\mathbf{z}_3$) different. Every patch is mapped to an instance prediction (colored by object, background patches in gray) by the instance projection $\phi$. The instance-level contrastive loss $\mathcal{L}_{\mathrm{instance}}$ clusters predictions of the same object together while keeping them separable from other objects and the background.}
  \vspace{-3mm}
  \label{fig:patchification}
\end{figure*}

Object-centric representation learning promises better data efficiency than image-level learning, because models can learn from multiple objects in one image. Image-level self-supervised methods align augmented views of the same image. But under random cropping, two views often capture different regions of a scene, potentially featuring different objects. A good encoder should arguably assign these regions different representations. However, image-level alignment forces these two views to agree and encourages the encoder to capture global, high-level semantics at every patch. 
As a result, each patch is incentivized to encode context rather than the object at its location, to the point that patches are repurposed to store global information~\citep{darcet2024vision}. Accordingly, recent models add dense losses to improve performance on downstream tasks that rely on localized information~\citep{oquab2024DINOv2, simeoni2025DINOv3}.
Object-level alignment instead removes the conflict between cross-view agreement and distinct representations for distinct objects.
Rather than matching scenes to scenes, it matches objects to objects: only representations of the same object need to agree across views, while distinct objects remain free to differ. This makes the representation modular, allowing the encoder to separate object identity from scene context.

Aligning object representations is more challenging than aligning whole images. While image-level alignment has clearly defined positive samples (augmented views of the same image), such a definition at the object level requires masks. Previous approaches to object-centric learning tried to solve partitioning and representation learning jointly in an end-to-end manner~\citep{eslami2016Attend, locatello2020ObjectCentric}. On natural images, this strategy has so far only worked when relying on frozen, large-scale pre-trained encoders to provide the patch features~\citep{seitzer2023Bridging}, which makes it difficult to test whether object-centric learning is more data-efficient. 
End-to-end approaches that do not rely on pre-trained encoders learn partitions that do not align with objects in crowded scenes \citep{henaff2022Object, lebailly2024cribo, wen2025DataCentric}. Their representations are then aligned across views per learned region rather than per object, so the encoder is never actually trained on objects and cannot show whether object-centric learning is more data-efficient.
In this work, we isolate the object representation learning problem and propose an approach that makes each image count for more, assuming masks are given, and assess whether it can attain higher data efficiency. 

Different downstream tasks have different demands on object-centric representations. Classification calls for a largely semantic space, in which the representations of two instances with the same semantics lie close together, e.g., two horses in the same image should have similar representations. Tracking, in contrast, calls for a more instance-specific space: watching two nearly identical horses race, we still need to tell them apart. These demands are in tension. Semantic similarity pulls instances of the same category together while instance-specificity pushes them apart. 
Encoders trained with image-level losses at large scale resolve this tension by depth: final-layer representations are largely semantic, while instance-specific object binding resides mainly in intermediate layers~\citep{li2026Does}. Scale improves both, but the instance-level structure of final-layer features lags far behind their semantics. We instead aim for a final-layer representation that holds both, while keeping them separable.

This work proposes an object-centric extension of LeJEPA in which alignment and the SIGReg anti-collapse regularizer act on the variable-sized set of objects in a scene rather than on whole images (see Figure~\ref{fig:patchification}). Object-level alignment, however, does not force the encoder to separate an object from its surroundings. We therefore add two components: copy-paste augmentation, which places objects in new contexts so that only the object itself is shared across views, and an instance-separating contrastive loss, which lets each patch predict the instance representation of its object and keeps co-occurring objects apart.
In our experimental assessment, we train on COCO fractions from 256 images to the full dataset and evaluate on ADE20k, DAVIS, NAVI and ImageNet-1k, using various downstream tasks. 
We further study different mask sources to understand whether benefits of the proposed approach persist with object masks from an off-the-shelf segmenter and with low-level Felzenszwalb--Huttenlocher segments~\citep{felzenszwalb2004efficient}.

\section{Related Work}
We place our work at the intersection of three lines of work: the objectives used to train visual encoders, the region-level pre-training methods that align image parts rather than whole images, and object-centric models that learn to decompose scenes and represent objects. Our setting is mask-supervised rather than self-supervised, so we borrow the objectives of self-supervised learning but not its label-free premise.

\subsection{Training Objectives}
Pixel-level reconstruction objectives, such as masked autoencoding \citep{he2022masked}, tend to produce features that are less informative for perception than alignment-based objectives \citep{van2026joint}. Those that aim to align representations across different augmented views of the same input differ in how they avoid the trivial solution of mapping every input to the same point. Contrastive losses~\citep{oord2019Representation, chen2020simple} employ repulsion from explicit negatives, which works when negatives are trustworthy but is unreliable when two samples may share their semantics~\citep{khosla2020Supervised}. Negative-free methods avoid this through teacher--student architectures, stop-gradients and schedules~\citep{caron2021Emerging, zhou2021IBOT, oquab2024DINOv2, simeoni2025DINOv3}. The LeJEPA framework~\citep{balestriero2025LeJEPA} replaces these heuristics with a principled anti-collapse objective, SIGReg, that regularizes the representations toward an isotropic Gaussian through a statistical test for normality. 
Rather than learning representations with either a contrastive or a negative-free objective, we project them into a \emph{semantic} and an \emph{instance space} and train each with the objective suited to it. The negative-free LeJEPA objective shapes the semantic space, since objects in a batch may share a category and would be false negatives, and a contrastive objective shapes the instance space, where every other object in the same image is a negative by construction.

\subsection{Region-level Representation Learning}
A growing body of work moves pre-training from whole images to patches, aligning corresponding local features across views instead of pooling the entire scene~\citep{wang2021Dense, bardes2022VICRegL}. Other works align representations at the coarser granularity of regions covering multiple patches. DetCon~\citep{henaff2021Efficient} applies a contrastive loss on mask-pooled features. Pooled representations from the same mask but different augmentations serve as positives. The remaining representations extracted in a batch serve as negatives.
CP$^2$~\citep{wang2022cp} also contrasts regions across copy-paste augmentations, but its regions are rectangles or other shapes that do not follow objects. Crops pasted onto other images are matched to their source, and a dense contrastive loss separates pasted from native pixels.
SlotCon~\citep{wen2022self} and its successor SlotMIM~\citep{wen2025DataCentric} pool features over the regions of a learned partition and train the pooled representations with a contrastive loss within a teacher--student setting.
In all of these, the only force shaping region representations is contrastive, so two semantically similar regions in the same image repel each other. We contrast only in the \emph{instance space}, where every other object in the image is a negative by construction, and paste objects rather than crops.
CrIBo~\citep{lebailly2024cribo} and OCEBO~\citep{djukic2025object} align object representations across views and avoid collapse through a teacher--student framework. As opposed to these methods we extend the LeJEPA framework to objects and propose a simple way to perform negative-free representation learning on objects without heuristics.
R-MAE~\citep{nguyen2024r} chooses a reconstruction-based approach and trains a decoder to complete masked binary region maps combined with traditional masked autoencoding. Unlike R-MAE, we learn through alignment rather than reconstruction.
\citet{mishra2022ObjectAware} incorporate regions in the augmentation pipeline and crop views around proposed objects rather than at random. Their encoder is restricted to train on object-centred views and is never explicitly encouraged to separate instances within a scene.

\subsection{Object-centric Learning}
Object-centric models aim to decompose a scene into objects and to represent each of them~\citep{eslami2016Attend, locatello2020ObjectCentric, wen2022self, lebailly2024cribo, djukic2025object, wen2025DataCentric}. 
On natural images, jointly learning partitioning and representation has not been achieved. Models that learn both from scratch produce partitions that do not align with objects (see Figure~\ref{fig:slotmim-partitions} in the appendix and the visualizations of \citealp{henaff2022Object, lebailly2024cribo}), which keeps them from learning object-centric representations in the first place~\citep{rubinstein2025Are}. Models that do partition well do so on top of frozen encoders trained on millions of images~\citep{seitzer2023Bridging, Didolkar2024ZeroShotOCRL}. We isolate representation learning from both partitioning and distillation from foundation models: we take masks as given and develop an object-centric representation framework aimed at learning more from few images.


\section{Method}
During training each image is augmented into several views, which the encoder $g$ maps independently to patch features. 
Given the masks, projected into each view, we extract two kinds of object representation from the patch features. An aggregator $f$ pools the patches of each object into one semantic object representation, which an object-centric LeJEPA loss $\mathcal{L}_{\mathrm{ObjectLeJEPA}}$ aligns across views. 
A head $\phi$ maps every patch to an object-binding patch representation, which a contrastive loss $\mathcal{L}_{\mathrm{instance}}$ makes agree within a mask and differ from all other patches.
This is visualized in Figure~\ref{fig:patchification}. Additional copy-paste augmentation, described last, moves objects between images so that only the object itself is shared across views. At inference time no masks are needed.

\subsection{Object Representation Extraction}
\label{sec:object_rep_extraction}
Each of the $B$ images in a batch yields $V$ views. $\mathcal{V}$ is the set of all $BV$ views and $g(\mathbf{x}_v)_i$ the feature of patch $i$ in view $v$. $\mathcal{K}$ is the set of objects in the batch, and each object's mask is projected into every view of its image. An object $k$ is \emph{present} in view $v$ if its projected mask covers more than $16\times16$ pixels (one ViT patch). $\mathcal{V}_k \subseteq \mathcal{V}$ is the set of views in which it is present, which, due to per-view cropping, need not be all views of its image. A patch is \emph{assigned} to object $k$ if the mask covers more than half of the patch. Unassigned patches are background. Views are produced with the geometric and photometric augmentations of LeJEPA~\citep{balestriero2025LeJEPA}, plus the copy-paste augmentation of Section~\ref{sec:copy_paste}.


\paragraph{Semantic Object Representations}
Following the principle that a whole is defined by its parts, we compute a semantic object representation $\mathbf{z}$ as a function $f$ of the patch features that fall within its mask. 
\begin{figure}[h]
  \centering
  \begin{tikzpicture}[
    font=\scriptsize,
    >=Stealth,
    box/.style={draw, rounded corners=2pt, minimum height=6mm, align=center, inner sep=2pt},
    io/.style={align=center, inner sep=1pt},
    add/.style={draw, circle, inner sep=0pt, minimum size=4mm},
    lbl/.style={font=\tiny, inner sep=1.2pt},
    node distance=5mm,
  ]
    \node[io] (gx) {$g(\mathbf{x})$};
    \node[box, right=of gx]    (mlp1)  {MLP};
    \node[box, right=of mlp1]  (wmean) {mask-\\weighted\\mean};
    \node[box, right=of wmean] (xattn) {masked\\cross\\attention};
    \node[add, right=of xattn] (add)   {$+$};
    \node[box, right=of add]   (mlp2)  {Residual\\MLP};
    \node[io,  right=of mlp2]  (z)     {$\mathbf{z}$};
    \draw[->] (gx)    -- (mlp1);
    \draw[->] (mlp1)  -- (wmean);
    \draw[->] (wmean) -- node[lbl,above] {q} coordinate[pos=0.55] (bxa) (xattn);
    \draw[->] (xattn) -- (add);
    \draw[->] (add)   -- (mlp2);
    \draw[->] (mlp2)  -- (z);
    \coordinate (kvT)  at ($(mlp1.north)+(0,5mm)$);
    \coordinate (kvT2) at (kvT -| xattn);
    \draw[->] (mlp1.north) -- (kvT) -- node[lbl,above] {k,v} (kvT2) -- (xattn.north);
    \coordinate (rB)  at ($(wmean.south)+(0,-5mm)$);
    \coordinate (rB2) at (rB -| add);
    \draw[->] (wmean) -- (wmean |- rB) -- node[lbl,below] {} (rB2) -- (add.south);
    \coordinate (r2T)  at ($(add.north)+(0,5mm)$);
    \coordinate (r2T2) at (r2T -| mlp2);
  \end{tikzpicture}
  \caption{Diagram of the patch aggregator $f$ outputting the semantic object representation $\mathbf{z}$ given patch features $g(\mathbf{x})$ and a mask.}
  \label{fig:aggregator}
  \vspace{-3mm}
  \end{figure}
As illustrated in Figure~\ref{fig:aggregator}, $f$ first projects each patch feature with an MLP and takes the mask-weighted mean, in which each patch is weighted by the fraction of its area the mask covers. This mean representation queries a cross-attention layer whose keys and values are restricted to patches inside the mask. A residual connection adds the mean representation back, and a residual MLP yields the representation of object $k$ in view $v$ as $\mathbf{z}_{v,k}$. The design mirrors how ViTs~\citep{dosovitskiy2021Image} pool a global representation into the \texttt{[cls]} token, with the mask-weighted mean standing in for an object-specific \texttt{[cls]} token. As is common, the loss acts on a $d$-dimensional MLP projection $\tilde{\mathbf{z}}_{v,k}$.


\paragraph{Instance-level Object Representations}
Here we rely on the principle that a part should know which whole it belongs to. The instance prediction of patch $i$ is an MLP $\phi$ on its feature, $\ell_2$-normalized,
\begin{align}
  \mathbf{y}_{v,i} = \frac{\phi(g(\mathbf{x}_{v})_i)}{\|\phi(g(\mathbf{x}_{v})_i)\|_2},
\end{align}
where $\mathbf{x}_{v}$ is view $v$ and $g(\mathbf{x}_{v})_i$ is the patch feature of the $i$-th patch.

\subsection{Object-centric LeJEPA}
The object-centric LeJEPA loss combines an alignment term and a regularization term. The alignment term $\mathcal{L}_{\mathrm{align}}$ encourages the representations of the same object across different views to be similar. We can only align when an object appears in at least two views. Hence, we define the effective set $\mathcal{K}^\mathrm{eff} = \{k \in \mathcal{K} \ | \ |\mathcal{V}_{k}| \geq 2\}$. We compute
\begin{align}
\ell^{\mathrm{align}}_{v,k} = \frac{1}{d} \| \boldsymbol{\mu}_{k} - \tilde{\mathbf{z}}_{v,k}\|_2^2, \ k \in \mathcal{K}^\mathrm{eff}, \ v \in \mathcal{V}_{k}
\end{align}
where $\boldsymbol{\mu}_{k} = \frac{1}{|\mathcal{V}_{k}|}\sum_{v \in \mathcal{V}_{k}} \tilde{\mathbf{z}}_{v,k}$ is the average of the $k$-th object representation across all views where it is present. The final alignment term $\mathcal{L}_{\mathrm{align}}$ is an average over all object-level alignment terms $\ell^{\mathrm{align}}_{v,k}$.
The regularization term $\mathcal{L}_{\mathrm{SIGReg}}$~\citep{balestriero2025LeJEPA} is applied on the projected object representations $\tilde{\mathbf{z}}_{v,k}$ to prevent collapse. Instead of calculating the empirical characteristic function with respect to representations of images, we compute it over all present object representations extracted from all $B$ images. In the original LeJEPA paper, SIGReg uses the Epps--Pulley test statistic, which is scaled by the sample size. In our case this sample size varies significantly across batches, since the number of objects in a view is far from constant. Hence, we scale the Epps--Pulley test statistic with a constant scale factor set to $B$ to avoid large fluctuations in gradient magnitude and unstable training.
The two terms are combined with weight $\lambda_{\mathrm{LeJEPA}} = 0.05$ as in LeJEPA,
\begin{align}
  \mathcal{L}_{\mathrm{ObjectLeJEPA}} = \mathcal{L}_{\mathrm{align}} + \lambda_{\mathrm{LeJEPA}}\, \mathcal{L}_{\mathrm{SIGReg}}.
\end{align}

\subsection{Instance-level Loss}
Alignment ties each object's representation across views, but nothing stops the features of neighbouring objects from leaking into each other, which is what makes co-occurring objects, such as the two horses of Figure~\ref{fig:patchification}, hard to tell apart. The instance loss makes every patch commit to its own object. It is computed within each view, since this is an intra-image problem, while cross-view consistency is handled by $\mathcal{L}_{\mathrm{align}}$. Following supervised contrastive learning~\citep{khosla2020Supervised}, we treat every assigned patch $i$ of view $v$ as an \emph{anchor}: $\mathcal{A}_v(i)$ is the set of all other patches of the view and $\mathcal{P}_v(i) \subseteq \mathcal{A}_v(i)$ those assigned to the same object. With temperature $\tau = 0.1$, the loss of an anchor with $|\mathcal{P}_v(i)| \geq 1$ is
\begin{align}
  \ell^{\mathrm{instance}}_{v,i} = -\frac{1}{|\mathcal{P}_v(i)|} \sum_{p \in \mathcal{P}_v(i)} \log \frac{\exp\!\left(\mathbf{y}_{v,i}^\top \mathbf{y}_{v,p} / \tau\right)}{\sum_{a \in \mathcal{A}_v(i)} \exp\!\left(\mathbf{y}_{v,i}^\top \mathbf{y}_{v,a} / \tau\right)}.
\end{align}
We average over anchors within a view, then over views. Background patches act as negatives only.
The total training objective is the unweighted sum of the two object-centric losses. We did not tune a relative weight:
\begin{align}
  \mathcal{L} = \mathcal{L}_{\mathrm{ObjectLeJEPA}} + \mathcal{L}_{\mathrm{instance}}.
\end{align}

\subsection{Copy-paste Augmentation}
\label{sec:copy_paste}
Alignment across crops and photometric changes of the same scene does not avoid the object representation from absorbing its context, because all views share it. We therefore paste objects into other images, as CP$^2$~\citep{wang2022cp} does with regions. Within a batch, images are paired as ``donor'' and ``host'' and one object from the donor is chosen per pair. From each cropped donor view the object is cut out with its mask, flipped, scaled, rotated, displaced from its donor position, and pasted onto one cropped host view. Photometric transforms follow. A pasted object counts as present in a host view if it meets the presence threshold, so $\mathcal{V}_k$ can span views from two images and $\boldsymbol{\mu}_k$ averages native and pasted appearances alike. A pasted copy shares nothing with its original but the object, so $\mathcal{L}_{\mathrm{align}}$ discourages encoding the scene. In the host view the pasted mask overlays the host's masks, and for $\mathcal{L}_{\mathrm{instance}}$ it is one more object. The composite view replaces the host's original view, and the donor's own views are training samples in any case, so copy-paste adds no encoded views and no forward passes. Sampling ranges and the transform details are in Appendix~\ref{sec:copy_paste_details}.

\section{Experiments and Results}
\label{sec:results}

\begin{table}[t]
\centering
\footnotesize
\setlength{\tabcolsep}{3pt}
\caption{Main results with frozen encoders (ViT-B/16; ViT-B/14 for DINOv2 and FT-DINOSAUR). Blocks group methods pre-trained on COCO with ground-truth annotations (top), on COCO without annotations (middle), and large-scale references (bottom). Bold marks the best value per block. Unless marked, instance-level and dense tasks use patch features, object-level tasks mask-averaged patch features, and image classification average-pooled patch features. For our method, object-level cells report mask-averaged patch features / semantic object representation $\mathbf{z}^{a}$. $^{b}$~\texttt{[cls]} token. Protocols in Section~\ref{sec:downstream_tasks}.}
\label{tab:main_results}
\resizebox{\textwidth}{!}{%
\begin{tabular}{@{}lrll@{\hskip 10pt}ccccccc@{}}
\toprule
\shortstack{{}\\{}\\\textbf{Method}} & \shortstack{{}\\{}\\\textbf{Ep}}& \shortstack{{}\\{}\\\textbf{Dataset}} & \shortstack{{}\\\textbf{GT}\\\textbf{source}} & \shortstack{\textbf{ADE20k}\\\textbf{Instance Clust.}\\\textbf{FG-ARI}} & \shortstack{\textbf{ADE20k}\\\textbf{Same-Class Pair}\\\textbf{AUC}} & \shortstack{\textbf{ADE20k}\\\textbf{Linear Seg.}\\\textbf{mIoU}} & \shortstack{\textbf{DAVIS}\\\textbf{Tracking}\\$\bm{\mathcal{J}\&\mathcal{F}}$} & \shortstack{\textbf{ADE20k}\\\textbf{Object Cls.}\\\textbf{Top-1 Bal. Acc.}} & \shortstack{\textbf{NAVI}\\\textbf{Re-ID}\\\textbf{5-shot Bal. Acc.}} & \shortstack{\textbf{ImageNet-1k}\\\textbf{Image Cls.}\\\textbf{Top-1 Bal. Acc.}} \\
\midrule
R-MAE & 2000 & COCO & panoptic & 0.343 & 0.840 & 0.374 & 0.576 & 0.484 & 0.623 & 0.488 \\
DetCon & 300 & COCO & instances & 0.303 & 0.822 & 0.375 & 0.621 & 0.495 & 0.564 & 0.513 \\
Crop-LeJEPA & 300 & COCO & boxes & 0.295 & 0.794 & 0.320 & 0.609 & 0.388 & 0.530 & 0.503$^{b}$ \\
Ours & 300 & COCO & instances & \textbf{0.423} & \textbf{0.850} & \textbf{0.403} & \textbf{0.683} & 0.531/\textbf{0.555}$^{a}$ & 0.714/\textbf{0.775}$^{a}$ & \textbf{0.574} \\
\midrule
MAE & 2000 & COCO & / & 0.336 & \textbf{0.836} & 0.365 & 0.577 & 0.466 & 0.644 & 0.479 \\
SlotMIM & 800 & COCO & / & \textbf{0.345} & \textbf{0.836} & \textbf{0.409} & \textbf{0.629} & \textbf{0.503} & 0.392 & \textbf{0.552} \\
Image LeJEPA & 300 & COCO & / & 0.298 & 0.801 & 0.343 & 0.626 & 0.426 & 0.609 & 0.472$^{b}$ \\
CP$^2$ & 400 & COCO & / & 0.268 & 0.702 & 0.281 & 0.574 & 0.337 & \textbf{0.737} & 0.454 \\
\midrule
FT-DINOSAUR &  & LVD-142M + COCO & / & \textbf{0.366} & 0.724 & 0.433 & 0.658 & 0.585 & 0.673 & 0.617 \\
DINOv2 &  & LVD-142M & / & 0.273 & 0.833 & 0.484 & 0.664 & \textbf{0.714} & 0.972 & 0.787$^{b}$ \\
DINOv3 &  & LVD-1689M & / & 0.354 & \textbf{0.871} & \textbf{0.500} & \textbf{0.715} & 0.703 & \textbf{0.986} & \textbf{0.790}$^{b}$ \\
\bottomrule
\end{tabular}
}
\vspace{-3mm}
\end{table}

We pre-train on COCO~\citep{lin2014Microsoft} with ground-truth instance masks and compare against three groups of baselines. The first block contains supervision-matched alternatives that also use COCO annotations during pre-training: DetCon~\citep{henaff2021Efficient} with instance masks, R-MAE~\citep{nguyen2024r} with panoptic masks, and Crop-LeJEPA with boxes, our variant of image-level LeJEPA whose views are crops from a dilated ground-truth box in the spirit of \citet{mishra2022ObjectAware}. The second block contains COCO-trained methods without annotations: MAE~\citep{he2022masked}, SlotMIM~\citep{wen2025DataCentric}, image-level LeJEPA~\citep{balestriero2025LeJEPA}, and CP$^2$~\citep{wang2022cp}. The third block lists large-scale references, DINOv2~\citep{oquab2024DINOv2}, DINOv3~\citep{simeoni2025DINOv3} and FT-DINOSAUR~\citep{Didolkar2024ZeroShotOCRL}, evaluated with their official checkpoints. Trained on $10^3$ to $10^4$ as many images, they are upper references rather than baselines. We train all other COCO baselines ourselves with the authors' code and recommended training schedules, with three deviations: CP$^2$ is trained from scratch rather than fine-tuned, DetCon and Crop-LeJEPA use a ViT backbone and the same multi-crop augmentations as the original LeJEPA, and R-MAE and MAE run for 2000 of the recommended 4000 epochs due to compute, still more than six times our 300 epochs. We train our model and image-level LeJEPA for $300$ epochs with the LeJEPA hyperparameters (AdamW, batch size $256$, $2$ global and $8$ local views). More training details can be found in Appendix~\ref{sec:training_details}.
Table~\ref{tab:main_results} groups the results by these three blocks.
Several other region-level methods only release ViT-S/16 or ResNet-50 checkpoints. We compare against these with a smaller backbone in Appendix~\ref{sec:smaller_archs}, where the conclusions are unchanged.

We freeze all encoders and evaluate them on a range of downstream tasks: linear probing for image classification on ImageNet~\citep{deng2009ImageNet}, various dense tasks on ADE20k~\citep{zhou2017Scene}, tracking via nearest neighbours on DAVIS~\citep{pont-tuset20182017}, and object re-identification from region-pooled patch features on NAVI~\citep{jampani2023NAVI}. Our downstream benchmarks span object types, scene layouts, and image distributions that differ from COCO, so they probe how well the frozen features generalize beyond the pre-training domain and object categories.


\subsection{Downstream Tasks}
\label{sec:downstream_tasks}
In the following paragraphs, we introduce downstream tasks and probes on frozen patch features and discuss the results. Further implementation details can be found in the appendix.
\paragraph{Instance-Awareness Probing}
To test whether encoders capture instance-level information, we first assign each patch to the mask that occupies the largest fraction of the patch. We discard patches assigned to the background. Then we apply two postprocessing pipelines to the frozen foreground patch features on ADE20k. First, we K-Means cluster the $\ell_2$-normalized features, setting the number of clusters to the number of ground-truth instance masks, and measure agreement with those masks via foreground adjusted rand index (FG-ARI). Second, following \citet{li2026Does}, we train a quadratic probe on all within-image pairs of foreground patches of the ADE20k training set to predict whether the two patches belong to the same instance. Since most pairs can be told apart from semantics alone, we evaluate only on pairs whose patches belong to objects of the same category and report the AUC: the probe is not asked whether a dog patch and a horse patch share an instance, but whether two horse patches do. Features encoding instance-level information should both cluster into object instances and support accurate same-instance prediction.

When clustering patch features into object instances (Table~\ref{tab:main_results}), our method leads every COCO-trained model with an FG-ARI of $0.423$ and also surpasses the large-scale references ($0.354$ for DINOv3, $0.366$ for FT-DINOSAUR). Two comparisons are telling. The end-to-end object-centric models, which learn their own partition, do not translate it into instance-aware patch features: SlotMIM reaches $0.345$ and FT-DINOSAUR, built on DINOv2, $0.366$. The supervision-matched alternatives receive the same masks as we do and still fall well short, DetCon at $0.303$ and R-MAE at $0.343$, so the gain comes from how the masks enter the objective, not from having them. The same-category quadratic probe tells a similar story: our method ($0.850$) is the best COCO-trained model and closes most of the gap to DINOv3 ($0.871$). 
Prior work~\citep{li2026Does} shows that the ability to tell whether two patches share an object emerges in pre-trained ViTs such as DINO, but that this signal may reside more strongly in intermediate layers than in the final one. Our instance-level loss pulls this object-membership information directly into the last-layer representation that downstream tasks consume.

\paragraph{Dense Semantic Tasks}
To test whether the encoders capture similarities across objects in different images, we evaluate the frozen patch features on two dense prediction tasks. First, we train a linear probe mapping frozen patch features to semantic segmentation labels on the ADE20k training set and evaluate it on the validation set, reporting mean IoU. Second, following the setup of \citet{caron2021Emerging}, we run a simple tracking pipeline on DAVIS in which the mask of the initial frame is propagated to subsequent frames by nearest-neighbour matching of the $\ell_2$-normalized frozen patch features, reporting the mean of contour accuracy and region similarity $\mathcal{J}\&\mathcal{F}$. Although this task also benefits from instance-level information, it requires matching objects across frames under appearance and pose changes.

On linear-probe semantic segmentation (Table~\ref{tab:main_results}), our method reaches $0.403$ mIoU, the best of the supervision-matched alternatives (DetCon $0.375$, R-MAE $0.374$) and on par with the best annotation-free model, SlotMIM at $0.409$ after 800 epochs. DINOv3 remains ahead at $0.500$. That SlotMIM matches us here is consistent with its design: its soft, semantic partition is well suited to category-level dense prediction, while it falls far behind on the instance-level tasks. On DAVIS label propagation, our method improves the $\mathcal{J}\&\mathcal{F}$ mean to $0.683$, ahead of every COCO-trained model in either block (best alternative: DetCon $0.621$, best annotation-free: SlotMIM $0.629$) and second only to DINOv3 ($0.715$). Our patch representations are therefore both temporally stable and object-discriminative. Notably, the instance-level loss only contrasts objects within a single view, yet the discriminability it induces together with the semantic loss transfers to separating objects across frames and scenes.
\paragraph{Object-level Tasks}
Because our model was trained with mask-guided object-level alignment, we evaluated whether the resulting object representations transfer to object-level downstream tasks. We consider two tasks. On ADE20k, we extracted a representation for each object using its ground-truth mask. For every method we averaged patch features within the mask and, additionally, computed the semantic object representations ($\mathbf{z}$) for our method. We trained a linear probe to predict the object class and computed top-1 balanced accuracy on the validation set. The second task was object re-identification on NAVI, which contains multiple images per object across varying backgrounds, poses, and lighting. Here we also extracted object representations using ground-truth masks and built a memory bank with 5 representations per object and classified each remaining image by its nearest neighbor in the bank, using $\ell_2$-normalized object representations. We repeat the memory-bank sampling 10 times and average. 

For linear classification of individual objects (Table~\ref{tab:main_results}), our semantic object representation $\mathbf{z}$ reaches $0.555$ balanced accuracy, ahead of the best supervision-matched alternative (DetCon, $0.495$) and the best annotation-free model (SlotMIM, $0.503$), though all COCO-trained models trail DINOv2 ($0.714$) and DINOv3 ($0.703$). On NAVI re-identification (5-shot), our semantic object representation $\mathbf{z}$ is the strongest among the COCO-trained models ($0.775$ against $0.564$ for DetCon and $0.737$ for CP$^2$). With mask-averaged patch features alone we reach $0.714$, second only to CP$^2$. The invariance induced by copy-paste appears to generalize to the pose and viewpoint changes of NAVI, underscored by the strong performance of CP$^2$ and our method. SlotMIM, in contrast, is the weakest annotation-free model ($0.392$), well below image-level LeJEPA ($0.609$). We attribute this to its soft slot assignments, which blend local and global information in the patch features.

\paragraph{Image-level Task}
We also evaluated image-level classification on ImageNet-1k. We trained a linear probe to predict the class label from the image representation and reported top-1 balanced accuracy on the validation set. For models without a \texttt{[cls]} token, we represented each image by globally average pooling its patch features.

Despite being trained to represent objects rather than whole images, our average-pooled patch features reach $0.574$ top-1 balanced accuracy, up from $0.472$ for image-level LeJEPA and ahead of the best supervision-matched alternative (DetCon, $0.513$) and the best annotation-free model (SlotMIM after 800 epochs, $0.552$). ImageNet is also the one benchmark on which Crop-LeJEPA clearly improves over image-level LeJEPA ($0.503$ vs.\ $0.472$) while trailing it on every other task: cropping views to single objects suits ImageNet's object-centred images, but discards the multi-object structure of COCO that mask-pooling exploits, and even on ImageNet it stays $7$ points behind our method. The appendix shows self-similarity maps on ImageNet images of categories unseen during training.
\subsection{Dataset Size Ablation}
\begin{figure}[t]
  \centering
  \begin{tikzpicture}
    \pgfplotsset{set layers}
    \pgfplotsset{
      every axis/.append style={
        width=0.28\textwidth, height=3.6cm,
        mark options={solid, draw opacity=0},
        xmode=log, log basis x=10,
        xtick={256,512,1024,11828,118287},
        xticklabels={256,512,1024,11.8k,118k},
        xticklabel style={rotate=45, anchor=north east, font=\tiny},
        minor tick num=0, log ticks with fixed point,
        enlarge x limits=0.12,
        grid=major, grid style={gray!20},
        tick align=outside,
        title style={align=center, font=\small},
      },
      imagelejepavitb/.style={color=objA, mark=*, thick},
      imagelejepavits/.style={color=objA, mark=square*, densely dashed, thick},
      detconvitb/.style={color=objC, mark=*, thick},
      detconvits/.style={color=objC, mark=square*, densely dashed, thick},
      oursvitb/.style={color=objB, mark=*, thick},
      oursvits/.style={color=objB, mark=square*, densely dashed, thick},
    }
    \begin{groupplot}[
        group style={group size=4 by 1, horizontal sep=0.95cm},
      ]
      \nextgroupplot[
          title={DAVIS $\mathcal{J}\&\mathcal{F}$},
          ymin=0.362, ymax=0.715,
        ]
        \pgfplotsonlayer{axis background}
        \fill[gray!12] (rel axis cs:0,0) rectangle ({axis cs:3480,0} |- {rel axis cs:0,1});
        \endpgfplotsonlayer
        \addplot[imagelejepavitb]  coordinates {(256,0.417) (512,0.441) (1024,0.461)};
        \addplot[imagelejepavitb, forget plot] coordinates {(11828,0.564) (118287,0.613)};
        \addplot[imagelejepavits]  coordinates {(256,0.386) (512,0.403) (1024,0.459)};
        \addplot[imagelejepavits, forget plot] coordinates {(11828,0.576) (118287,0.608)};
        \addplot[detconvitb]  coordinates {(256,0.493) (512,0.529) (1024,0.549)};
        \addplot[detconvitb, forget plot] coordinates {(11828,0.601) (118287,0.627)};
        \addplot[detconvits]  coordinates {(256,0.496) (512,0.526) (1024,0.554)};
        \addplot[detconvits, forget plot] coordinates {(11828,0.601) (118287,0.613)};
        \addplot[oursvitb]  coordinates {(256,0.633) (512,0.632) (1024,0.646)};
        \addplot[oursvitb, forget plot] coordinates {(11828,0.659) (118287,0.691)};
        \addplot[oursvits]  coordinates {(256,0.633) (512,0.632) (1024,0.639)};
        \addplot[oursvits, forget plot] coordinates {(11828,0.652) (118287,0.676)};

      \nextgroupplot[
          title={ImageNet-1k top-1},
          ymin=0.111, ymax=0.606,
        ]
        \pgfplotsonlayer{axis background}
        \fill[gray!12] (rel axis cs:0,0) rectangle ({axis cs:3480,0} |- {rel axis cs:0,1});
        \endpgfplotsonlayer
        \addplot[imagelejepavitb]  coordinates {(256,0.180) (512,0.186) (1024,0.201)};
        \addplot[imagelejepavitb, forget plot] coordinates {(11828,0.334) (118287,0.473)};
        \addplot[imagelejepavits]  coordinates {(256,0.145) (512,0.146) (1024,0.165)};
        \addplot[imagelejepavits, forget plot] coordinates {(11828,0.269) (118287,0.436)};
        \addplot[detconvitb]  coordinates {(256,0.279) (512,0.298) (1024,0.320)};
        \addplot[detconvitb, forget plot] coordinates {(11828,0.435) (118287,0.516)};
        \addplot[detconvits]  coordinates {(256,0.251) (512,0.267) (1024,0.285)};
        \addplot[detconvits, forget plot] coordinates {(11828,0.390) (118287,0.451)};
        \addplot[oursvitb]  coordinates {(256,0.345) (512,0.360) (1024,0.373)};
        \addplot[oursvitb, forget plot] coordinates {(11828,0.483) (118287,0.572)};
        \addplot[oursvits]  coordinates {(256,0.286) (512,0.298) (1024,0.316)};
        \addplot[oursvits, forget plot] coordinates {(11828,0.428) (118287,0.512)};

      \nextgroupplot[
          title={ADE20k mIoU},
          ymin=0.189, ymax=0.416,
        ]
        \pgfplotsonlayer{axis background}
        \fill[gray!12] (rel axis cs:0,0) rectangle ({axis cs:3480,0} |- {rel axis cs:0,1});
        \endpgfplotsonlayer
        \addplot[imagelejepavitb]  coordinates {(256,0.220) (512,0.226) (1024,0.232)};
        \addplot[imagelejepavitb, forget plot] coordinates {(11828,0.283) (118287,0.339)};
        \addplot[imagelejepavits]  coordinates {(256,0.204) (512,0.208) (1024,0.217)};
        \addplot[imagelejepavits, forget plot] coordinates {(11828,0.271) (118287,0.319)};
        \addplot[detconvitb]  coordinates {(256,0.244) (512,0.256) (1024,0.268)};
        \addplot[detconvitb, forget plot] coordinates {(11828,0.337) (118287,0.375)};
        \addplot[detconvits]  coordinates {(256,0.237) (512,0.246) (1024,0.258)};
        \addplot[detconvits, forget plot] coordinates {(11828,0.322) (118287,0.348)};
        \addplot[oursvitb]  coordinates {(256,0.305) (512,0.316) (1024,0.325)};
        \addplot[oursvitb, forget plot] coordinates {(11828,0.368) (118287,0.401)};
        \addplot[oursvits]  coordinates {(256,0.287) (512,0.296) (1024,0.304)};
        \addplot[oursvits, forget plot] coordinates {(11828,0.352) (118287,0.385)};

      \nextgroupplot[
          title={NAVI 5-shot},
          ymin=0.277, ymax=0.821,
        ]
        \pgfplotsonlayer{axis background}
        \fill[gray!12] (rel axis cs:0,0) rectangle ({axis cs:3480,0} |- {rel axis cs:0,1});
        \endpgfplotsonlayer
        \addplot[imagelejepavitb]  coordinates {(256,0.392) (512,0.365) (1024,0.369)};
        \addplot[imagelejepavitb, forget plot] coordinates {(11828,0.492) (118287,0.556)};
        \addplot[imagelejepavits]  coordinates {(256,0.407) (512,0.405) (1024,0.434)};
        \addplot[imagelejepavits, forget plot] coordinates {(11828,0.525) (118287,0.561)};
        \addplot[detconvitb]  coordinates {(256,0.340) (512,0.336) (1024,0.340)};
        \addplot[detconvitb, forget plot] coordinates {(11828,0.417) (118287,0.578)};
        \addplot[detconvits]  coordinates {(256,0.326) (512,0.317) (1024,0.315)};
        \addplot[detconvits, forget plot] coordinates {(11828,0.418) (118287,0.572)};
        \addplot[oursvitb]  coordinates {(256,0.713) (512,0.743) (1024,0.737)};
        \addplot[oursvitb, forget plot] coordinates {(11828,0.651) (118287,0.783)};
        \addplot[oursvits]  coordinates {(256,0.695) (512,0.688) (1024,0.693)};
        \addplot[oursvits, forget plot] coordinates {(11828,0.650) (118287,0.778)};

    \end{groupplot}
    \coordinate (xcenter) at ($(group c2r1.south east)!0.5!(group c3r1.south west)$);
    \node[anchor=north, yshift=-2pt] at (xcenter |- current bounding box.south)
      {Pre-training images (COCO subset size)\quad{\footnotesize\color{gray!70}shaded: 15k steps $\cdot$ unshaded: $\sim$69k}};
    \node[anchor=south, yshift=2pt, font=\footnotesize] at (xcenter |- current bounding box.north)
      {\tikz[baseline=-0.55ex]{\draw[objA, thick] (0,0)--(0.45,0); \fill[objA] (0.23,0) circle(1.7pt);}\,Image LeJEPA\hspace{1.3em}\tikz[baseline=-0.55ex]{\draw[objC, thick] (0,0)--(0.45,0); \fill[objC] (0.23,0) circle(1.7pt);}\,DetCon\hspace{1.3em}\tikz[baseline=-0.55ex]{\draw[objB, thick] (0,0)--(0.45,0); \fill[objB] (0.23,0) circle(1.7pt);}\,Ours\hspace{2.4em}\tikz[baseline=-0.55ex]{\draw[black!70, thick] (0,0)--(0.45,0); \fill[black!70] (0.23,0) circle(1.7pt);}\,ViT-B\hspace{1.3em}\tikz[baseline=-0.55ex]{\draw[black!70, thick, densely dashed] (0,0)--(0.45,0); \fill[black!70] (0.17,-0.06) rectangle (0.29,0.06);}\,ViT-S};
  \end{tikzpicture}
  \caption{Dataset-size ablation across four downstream tasks, described in Section~\ref{sec:downstream_tasks}. The $x$-axis is the number of COCO images used for pre-training ($256$, $512$, $1024$, $11.8\text{k}\approx10\%$, $118\text{k}\approx100\%$). Each line is one (method, backbone) pair: Image LeJEPA (orange), DetCon (green), our method (blue), ViT-Base (solid, circles) vs. ViT-Small (dashed, squares). From left to right we report DAVIS tracking $\mathcal{J}\&\mathcal{F}$ on $\ell_2$-normalized patch features, ImageNet-1k logistic-regression top-1 balanced accuracy (\texttt{[cls]}-token for Image LeJEPA and average-pooled patch features for others), ADE20k linear-probe segmentation mIoU on patch features, and NAVI 5-shot balanced accuracy on the mask-pooled patch features or semantic object representation $\mathbf{z}$ for our method.}
  \label{fig:dataset-size-ablation}
  \vspace{-3mm}
\end{figure}
Figure~\ref{fig:dataset-size-ablation} presents results on our ablation study on the pre-training dataset size and the backbone capacity across the four downstream tasks of Section~\ref{sec:downstream_tasks}.
The $10\%$ and $100\%$ subsets train for about $69$k steps, same as $150$ epochs of full COCO. Subsets of at most $1024$ images train for a fixed budget of $15$k steps, which already amounts to $15$k epochs on $256$ images and $3.7$k on $1024$. The shaded and unshaded halves of each panel therefore differ in compute, and only points within one half are directly comparable.
Given masks, object-level alignment needs very few images for the instance-level tasks: with $1024$ images our ViT-B model already surpasses image-level LeJEPA trained on the full dataset on tracking ($0.646$ vs.\ $0.613$) and re-identification ($0.737$ vs.\ $0.556$). The semantic tasks benefit more from a larger corpus, but at $10\%$ of COCO our model exceeds full-data image-level LeJEPA on all four tasks (ImageNet $0.483$ vs.\ $0.473$, ADE20k $0.368$ vs.\ $0.339$). Object-level alignment also outweighs backbone capacity: our ViT-S model trained on full COCO beats image-level LeJEPA on ViT-B on every task. DetCon, the supervision-matched alternative, lies between the two everywhere except on NAVI, where it stays below image-level LeJEPA up to $10\%$ of COCO ($0.417$ vs.\ $0.492$ at $11.8$k images) and only overtakes it on the full dataset.

The dip on NAVI between $1024$ and $11.8$k images reflects a change in which loss mostly guides the encoder. On tiny datasets cross-view agreement for a few hundred objects is quickly memorized, so the alignment term stops providing gradients and instance separation alone shapes the encoder, which favours re-identification. With more data alignment becomes informative again: the semantic tasks jump and re-identification dips, until at full scale the encoder captures both semantic and instance-level information. Appendix~\ref{sec:regime_transition} provides more details.

\subsection{Component \& Mask Ablation}
\label{sec:component_ablation}
\begin{table}[t]
\centering
\footnotesize
\setlength{\tabcolsep}{3pt}
\caption{Ablation of our losses, copy-paste augmentations and mask source on the full COCO dataset with a ViT-B backbone, trained for $150$ epochs. We report the same downstream tasks as Table~\ref{tab:main_results}. In the top block each row adds one component to the row above. Within the top block the best value per column is marked bold. The bottom section shows results for swapping COCO GT masks with SAM masks or Felzenszwalb--Huttenlocher (FH) superpixels. Readouts as in Table~\ref{tab:main_results}: $^{a}$~semantic object representation $\mathbf{z}$, $^{b}$~\texttt{[cls]} token.}
\label{tab:ablation}
\resizebox{\textwidth}{!}{%
\begin{tabular}{@{}ll@{\hskip 10pt}ccccccc@{}}
\toprule
\shortstack{{}\\{}\\\textbf{Method}} & \shortstack{{}\\{}\\\textbf{Mask source}} & \shortstack{\textbf{ADE20k}\\\textbf{Instance Clust.}\\\textbf{FG-ARI}} & \shortstack{\textbf{ADE20k}\\\textbf{Same-Class Pair}\\\textbf{AUC}} & \shortstack{\textbf{ADE20k}\\\textbf{Linear Seg.}\\\textbf{mIoU}} & \shortstack{\textbf{DAVIS}\\\textbf{Tracking}\\$\bm{\mathcal{J}\&\mathcal{F}}$} & \shortstack{\textbf{ADE20k}\\\textbf{Object Cls.}\\\textbf{Top-1 Bal. Acc.}} & \shortstack{\textbf{NAVI}\\\textbf{Re-ID}\\\textbf{5-shot Bal. Acc.}} & \shortstack{\textbf{ImageNet-1k}\\\textbf{Image Cls.}\\\textbf{Top-1 Bal. Acc.}} \\
\midrule
Image LeJEPA & / & 0.287 & 0.802 & 0.339 & 0.613 & 0.419 & 0.556 & 0.473$^{b}$ \\
$\mathcal{L}_{\mathrm{ObjectLeJEPA}}$ & GT instances & 0.312 & 0.822 & 0.373 & 0.635 & 0.522$^{a}$ & 0.512$^{a}$ & 0.525 \\
+ $\mathcal{L}_{\mathrm{instance}}$ & GT instances & 0.409 & \textbf{0.851} & 0.399 & 0.669 & 0.536$^{a}$ & 0.562$^{a}$ & 0.554 \\
+ Copy-Paste (Ours) & GT instances & \textbf{0.422} & \textbf{0.851} & \textbf{0.401} & \textbf{0.691} & \textbf{0.563}$^{a}$ & \textbf{0.783}$^{a}$ & \textbf{0.572} \\
\midrule
Ours & SAM & 0.424 & 0.890 & 0.424 & 0.691 & 0.582$^{a}$ & 0.667$^{a}$ & 0.542 \\
Ours & FH superpixels & 0.381 & 0.845 & 0.370 & 0.605 & 0.512$^{a}$ & 0.676$^{a}$ & 0.483 \\
\bottomrule
\end{tabular}
}
\vspace{-3mm}
\end{table}

Table~\ref{tab:ablation} presents the contribution of each component for the different tasks. Switching from image-level to object-level alignment alone improves every task except re-identification, where NAVI drops from $0.556$ to $0.512$. Aligning an object across augmented views of the same scene does not force the encoder to separate the object from its surroundings. The object representation still absorbs scene context, which helps category-level tasks (object classification $0.419 \to 0.522$) but hurts re-identification across scenes. The instance loss adds the instance-level structure that semantic alignment alone lacks (FG-ARI $0.312 \to 0.409$, DAVIS $0.635 \to 0.669$) and copy-paste closes the remaining gap. It improves every task, with the largest gain on NAVI ($0.562 \to 0.783$). Copy-paste trains the encoder to recognize an object in a new background, and this generalizes to recognizing it in a new pose. Only the same-class AUC is saturated at $0.851$.

The bottom block replaces the ground-truth masks with SAM-generated masks and FH superpixels. SAM masks match or exceed ground truth on the instance-level, dense semantic tasks and object classification but fall behind on re-identification and image classification. The benefit of object-level alignment thus does not hinge on perfect annotations. FH superpixels are not object masks, yet our method still outperforms image-level LeJEPA on most tasks with them, which shows its robustness to the partition.

\section{Conclusion and Limitations}
Object-level alignment can extract more from each image than image-level alignment, but only once objects are kept apart from their context. We extend LeJEPA from images to objects, so that alignment and the SIGReg regularizer act on the set of objects in a batch. Two additions keep object representations from absorbing their surroundings, namely copy-paste augmentation and an instance-separating contrastive loss in a separate projection space. Given the same images, our method outperforms image-level LeJEPA and a supervision-matched alternative that uses the same masks, at every data scale from $256$ images to full COCO. The benefit persists with masks from an off-the-shelf segmenter, so the training images need not be manually annotated. We hope that the losses and augmentation developed here with given masks carry over to end-to-end models that discover their own partitions.

Our results hold for given masks and COCO-scale training. The mask sources are not label-free, as ground-truth masks are human-annotated and SAM was itself trained on annotated data. Each configuration is trained with a single seed. Our method is, however, the best of the three methods in all ten settings of Figure~\ref{fig:dataset-size-ablation}, across two backbones and five data scales.
Behavior at larger data and model scales remains untested.

\section*{Reproducibility Statement}
All datasets we use (COCO, ADE20k, DAVIS, NAVI, ImageNet-1k) are publicly available. Training hyperparameters are given in Appendix~\ref{sec:training_details}, the copy-paste augmentation in Appendix~\ref{sec:copy_paste_details}, and the protocols and probes for every downstream task in Appendix~\ref{sec:downstream_details}. Baselines were evaluated with their official checkpoints where available and otherwise trained with the authors' public code, with the deviations stated in Section~\ref{sec:results}. In particular, DetCon uses a ViT backbone and the multi-crop augmentations of Image LeJEPA and our method for comparability. Code to train our model and to generate the SAM masks will be released publicly.


\bibliography{iclr2027_conference}

\begin{thebibliography}{38}
\providecommand{\natexlab}[1]{#1}
\providecommand{\url}[1]{\texttt{#1}}
\expandafter\ifx\csname urlstyle\endcsname\relax
  \providecommand{\doi}[1]{doi: #1}\else
  \providecommand{\doi}{doi: \begingroup \urlstyle{rm}\Url}\fi

\bibitem[Balestriero \& LeCun(2025)Balestriero and LeCun]{balestriero2025LeJEPA}
Randall Balestriero and Yann LeCun.
\newblock {{LeJEPA}}: {{Provable}} and {{Scalable Self-Supervised Learning Without}} the {{Heuristics}}, November 2025.

\bibitem[Bardes et~al.(2022)Bardes, Ponce, and LeCun]{bardes2022VICRegL}
Adrien Bardes, Jean Ponce, and Yann LeCun.
\newblock {{VICRegL}}: {{Self-Supervised Learning}} of {{Local Visual Features}}.
\newblock \emph{Advances in Neural Information Processing Systems}, 35:\penalty0 8799--8810, December 2022.

\bibitem[Caron et~al.(2021)Caron, Touvron, Misra, J{\'e}gou, Mairal, Bojanowski, and Joulin]{caron2021Emerging}
Mathilde Caron, Hugo Touvron, Ishan Misra, Herv{\'e} J{\'e}gou, Julien Mairal, Piotr Bojanowski, and Armand Joulin.
\newblock Emerging properties in self-supervised vision transformers.
\newblock In \emph{Proceedings of the {{IEEE}}/{{CVF}} International Conference on Computer Vision}, pp.\  9650--9660, 2021.

\bibitem[Chen et~al.(2020)Chen, Kornblith, Norouzi, and Hinton]{chen2020simple}
Ting Chen, Simon Kornblith, Mohammad Norouzi, and Geoffrey Hinton.
\newblock A simple framework for contrastive learning of visual representations.
\newblock In \emph{International Conference on Machine Learning}, pp.\  1597--1607. PmLR, 2020.

\bibitem[Darcet et~al.(2024)Darcet, Oquab, Mairal, and Bojanowski]{darcet2024vision}
Timoth{\'e}e Darcet, Maxime Oquab, Julien Mairal, and Piotr Bojanowski.
\newblock Vision transformers need registers.
\newblock In \emph{International Conference on Learning Representations}, volume 2024, pp.\  2632--2652, 2024.

\bibitem[Deng et~al.(2009)Deng, Dong, Socher, Li, Li, and {Fei-Fei}]{deng2009ImageNet}
Jia Deng, Wei Dong, Richard Socher, Li-Jia Li, Kai Li, and Li~{Fei-Fei}.
\newblock {{ImageNet}}: {{A}} large-scale hierarchical image database.
\newblock In \emph{2009 {{IEEE Conference}} on {{Computer Vision}} and {{Pattern Recognition}}}, pp.\  248--255, June 2009.
\newblock \doi{10.1109/CVPR.2009.5206848}.

\bibitem[Didolkar et~al.(2024)Didolkar, Zadaianchuk, Goyal, Mozer, Bengio, Martius, and Seitzer]{Didolkar2024ZeroShotOCRL}
Aniket Didolkar, Andrii Zadaianchuk, Anirudh Goyal, Mike Mozer, Yoshua Bengio, Georg Martius, and Maximilian Seitzer.
\newblock Zero-shot object-centric representation learning.
\newblock \emph{arXiv:2408.09162}, 2024.

\bibitem[Dosovitskiy et~al.(2021)Dosovitskiy, Beyer, Kolesnikov, Weissenborn, Zhai, Unterthiner, Dehghani, Minderer, Heigold, Gelly, Uszkoreit, and Houlsby]{dosovitskiy2021Image}
Alexey Dosovitskiy, Lucas Beyer, Alexander Kolesnikov, Dirk Weissenborn, Xiaohua Zhai, Thomas Unterthiner, Mostafa Dehghani, Matthias Minderer, Georg Heigold, Sylvain Gelly, Jakob Uszkoreit, and Neil Houlsby.
\newblock An {{Image}} is {{Worth}} 16x16 {{Words}}: {{Transformers}} for {{Image Recognition}} at {{Scale}}.
\newblock In \emph{9th {{International Conference}} on {{Learning Representations}}, {{ICLR}} 2021, {{Virtual Event}}, {{Austria}}, {{May}} 3-7, 2021}. OpenReview.net, 2021.

\bibitem[Eslami et~al.(2016)Eslami, Heess, Weber, Tassa, Szepesvari, {kavukcuoglu}, and Hinton]{eslami2016Attend}
S.~M.~Ali Eslami, Nicolas Heess, Theophane Weber, Yuval Tassa, David Szepesvari, koray {kavukcuoglu}, and Geoffrey~E Hinton.
\newblock Attend, infer, repeat: {{Fast}} scene understanding with generative models.
\newblock In D.~D. Lee, M.~Sugiyama, U.~V. Luxburg, I.~Guyon, and R.~Garnett (eds.), \emph{Advances in Neural Information Processing Systems 29}, pp.\  3225--3233. Curran Associates, Inc., 2016.

\bibitem[Felzenszwalb \& Huttenlocher(2004)Felzenszwalb and Huttenlocher]{felzenszwalb2004efficient}
Pedro~F Felzenszwalb and Daniel~P Huttenlocher.
\newblock Efficient graph-based image segmentation.
\newblock \emph{International journal of computer vision}, 59\penalty0 (2):\penalty0 167--181, 2004.

\bibitem[He et~al.(2022)He, Chen, Xie, Li, Doll{\'a}r, and Girshick]{he2022masked}
Kaiming He, Xinlei Chen, Saining Xie, Yanghao Li, Piotr Doll{\'a}r, and Ross Girshick.
\newblock Masked autoencoders are scalable vision learners.
\newblock In \emph{2022 {{IEEE}}/{{CVF}} Conference on Computer Vision and Pattern Recognition ({{CVPR}})}, pp.\  15979--15988. IEEE, 2022.

\bibitem[H{\'e}naff et~al.(2021)H{\'e}naff, Koppula, Alayrac, van~den Oord, Vinyals, and Carreira]{henaff2021Efficient}
Olivier~J H{\'e}naff, Skanda Koppula, Jean-Baptiste Alayrac, Aaron van~den Oord, Oriol Vinyals, and Jo{\~a}o Carreira.
\newblock Efficient visual pretraining with contrastive detection.
\newblock \emph{International Conference on Computer Vision}, 2021.

\bibitem[H{\'e}naff et~al.(2022)H{\'e}naff, Koppula, Shelhamer, Zoran, Jaegle, Zisserman, Carreira, and Arandjelovi{\'c}]{henaff2022Object}
Olivier~J H{\'e}naff, Skanda Koppula, Evan Shelhamer, Daniel Zoran, Andrew Jaegle, Andrew Zisserman, Jo{\~a}o Carreira, and Relja Arandjelovi{\'c}.
\newblock Object discovery and representation networks.
\newblock In \emph{European Conference on Computer Vision}, pp.\  123--143. Springer, 2022.

\bibitem[Jampani et~al.(2023)Jampani, Maninis, Engelhardt, Karpur, Truong, Sargent, Popov, Araujo, {Martin-Brualla}, Patel, Vlasic, Ferrari, Makadia, Liu, Li, and Zhou]{jampani2023NAVI}
Varun Jampani, Kevis-Kokitsi Maninis, Andreas Engelhardt, Arjun Karpur, Karen Truong, Kyle Sargent, Stefan Popov, Andre Araujo, Ricardo {Martin-Brualla}, Kaushal Patel, Daniel Vlasic, Vittorio Ferrari, Ameesh Makadia, Ce~Liu, Yuanzhen Li, and Howard Zhou.
\newblock {{NAVI}}: {{Category-agnostic}} image collections with high-quality {{3D}} shape and pose annotations.
\newblock In \emph{{{NeurIPS}}}, 2023.

\bibitem[Khosla et~al.(2020)Khosla, Teterwak, Wang, Sarna, Tian, Isola, Maschinot, Liu, and Krishnan]{khosla2020Supervised}
Prannay Khosla, Piotr Teterwak, Chen Wang, Aaron Sarna, Yonglong Tian, Phillip Isola, Aaron Maschinot, Ce~Liu, and Dilip Krishnan.
\newblock Supervised contrastive learning.
\newblock \emph{Advances in neural information processing systems}, 33:\penalty0 18661--18673, 2020.

\bibitem[Lebailly et~al.(2024)Lebailly, Stegm{\"u}ller, Bozorgtabar, Thiran, and Tuytelaars]{lebailly2024cribo}
Tim Lebailly, Thomas Stegm{\"u}ller, Behzad Bozorgtabar, Jean-Philippe Thiran, and Tinne Tuytelaars.
\newblock Cribo: {{Self-supervised}} learning via cross-image object-level bootstrapping.
\newblock In \emph{International Conference on Learning Representations}, volume 2024, pp.\  46794--46814, 2024.

\bibitem[Li et~al.(2026)Li, Salehi, Ungar, and Kording]{li2026Does}
Yihao Li, Saeed Salehi, Lyle Ungar, and Konrad Kording.
\newblock Does object binding naturally emerge in large pretrained vision transformers?
\newblock \emph{Advances in Neural Information Processing Systems}, 38:\penalty0 3394--3423, 2026.

\bibitem[Lin et~al.(2014)Lin, Maire, Belongie, Hays, Perona, Ramanan, Doll{\'a}r, and Zitnick]{lin2014Microsoft}
Tsung-Yi Lin, Michael Maire, Serge Belongie, James Hays, Pietro Perona, Deva Ramanan, Piotr Doll{\'a}r, and C~Lawrence Zitnick.
\newblock Microsoft coco: {{Common}} objects in context.
\newblock In \emph{European Conference on Computer Vision}, pp.\  740--755. Springer, 2014.

\bibitem[Locatello et~al.(2020)Locatello, Weissenborn, Unterthiner, Mahendran, Heigold, Uszkoreit, Dosovitskiy, and Kipf]{locatello2020ObjectCentric}
Francesco Locatello, Dirk Weissenborn, Thomas Unterthiner, Aravindh Mahendran, Georg Heigold, Jakob Uszkoreit, Alexey Dosovitskiy, and Thomas Kipf.
\newblock Object-{{Centric Learning}} with {{Slot Attention}}.
\newblock In \emph{Advances in {{Neural Information Processing Systems}}}, volume~33, pp.\  11525--11538. Curran Associates, Inc., 2020.
\newblock \doi{10.48550/arXiv.2006.15055}.

\bibitem[Mishra et~al.(2022)Mishra, Shah, Bansal, Anjaria, Jagannatha, Sharma, Jacobs, and Krishnan]{mishra2022ObjectAware}
Shlok~Kumar Mishra, Anshul Shah, Ankan Bansal, Janit~K Anjaria, Abhyuday~Narayan Jagannatha, Abhishek Sharma, David Jacobs, and Dilip Krishnan.
\newblock Object-aware cropping for self-supervised learning.
\newblock \emph{Transactions on Machine Learning Research}, 2022.
\newblock ISSN 2835-8856.

\bibitem[Nguyen et~al.(2024)Nguyen, Li, Aggarwal, Oswald, Kirillov, Snoek, and Chen]{nguyen2024r}
Duy-Kien Nguyen, Yanghao Li, Vaibhav Aggarwal, Martin~R Oswald, Alexander Kirillov, Cees~G Snoek, and Xinlei Chen.
\newblock R-{{MAE}}: {{Regions}} meet masked autoencoders.
\newblock In \emph{International Conference on Learning Representations}, volume 2024, pp.\  47519--47536, 2024.

\bibitem[OpenAI(2026)]{chatgpt2026}
OpenAI.
\newblock {{ChatGPT}}, 2026.

\bibitem[Oquab et~al.(2024)Oquab, Darcet, Moutakanni, Vo, Szafraniec, Khalidov, Fernandez, Haziza, Massa, {El-Nouby}, Assran, Ballas, Galuba, Howes, Huang, Li, Misra, Rabbat, Sharma, Synnaeve, Xu, Jegou, Mairal, Labatut, Joulin, and Bojanowski]{oquab2024DINOv2}
Maxime Oquab, Timoth{\'e}e Darcet, Th{\'e}o Moutakanni, Huy Vo, Marc Szafraniec, Vasil Khalidov, Pierre Fernandez, Daniel Haziza, Francisco Massa, Alaaeldin {El-Nouby}, Mahmoud Assran, Nicolas Ballas, Wojciech Galuba, Russell Howes, Po-Yao Huang, Shang-Wen Li, Ishan Misra, Michael Rabbat, Vasu Sharma, Gabriel Synnaeve, Hu~Xu, Herv{\'e} Jegou, Julien Mairal, Patrick Labatut, Armand Joulin, and Piotr Bojanowski.
\newblock {{DINOv2}}: {{Learning Robust Visual Features}} without {{Supervision}}, February 2024.

\bibitem[{Pont-Tuset} et~al.(2017){Pont-Tuset}, Perazzi, Caelles, Arbel{\'a}ez, {Sorkine-Hornung}, and Gool]{pont-tuset20182017}
Jordi {Pont-Tuset}, Federico Perazzi, Sergi Caelles, Pablo Arbel{\'a}ez, Alex {Sorkine-Hornung}, and Luc~Van Gool.
\newblock The 2017 {{DAVIS Challenge}} on {{Video Object Segmentation}}, 2017.

\bibitem[Ravi et~al.(2025)Ravi, Gabeur, Hu, Hu, Ryali, Ma, Khedr, R{\"a}dle, Rolland, Gustafson, et~al.]{ravi2025SAM}
Nikhila Ravi, Valentin Gabeur, Yuan-Ting Hu, Ronghang Hu, Chaitanya Ryali, Tengyu Ma, Haitham Khedr, Roman R{\"a}dle, Chloe Rolland, Laura Gustafson, et~al.
\newblock Sam 2: {{Segment}} anything in images and videos.
\newblock In \emph{International Conference on Learning Representations}, volume 2025, pp.\  28085--28128, 2025.

\bibitem[Rubinstein et~al.(2025)Rubinstein, Prabhu, Bethge, and Oh]{rubinstein2025Are}
Alexander Rubinstein, Ameya Prabhu, Matthias Bethge, and Seong~Joon Oh.
\newblock Are {{We Done}} with {{Object-Centric Learning}}?, April 2025.

\bibitem[Seitzer et~al.(2023)Seitzer, Horn, Zadaianchuk, Zietlow, Xiao, {Simon-Gabriel}, He, Zhang, Sch{\"o}lkopf, Brox, and Locatello]{seitzer2023Bridging}
Maximilian Seitzer, Max Horn, Andrii Zadaianchuk, Dominik Zietlow, Tianjun Xiao, Carl-Johann {Simon-Gabriel}, Tong He, Zheng Zhang, Bernhard Sch{\"o}lkopf, Thomas Brox, and Francesco Locatello.
\newblock Bridging the gap to real-world object-centric learning.
\newblock In \emph{The Eleventh International Conference on Learning Representations}, 2023.

\bibitem[Sim{\'e}oni et~al.(2025)Sim{\'e}oni, Vo, Seitzer, Baldassarre, Oquab, Jose, Khalidov, Szafraniec, Yi, Ramamonjisoa, Massa, Haziza, Wehrstedt, Wang, Darcet, Moutakanni, Sentana, Roberts, Vedaldi, Tolan, Brandt, Couprie, Mairal, J{\'e}gou, Labatut, and Bojanowski]{simeoni2025DINOv3}
Oriane Sim{\'e}oni, Huy~V. Vo, Maximilian Seitzer, Federico Baldassarre, Maxime Oquab, Cijo Jose, Vasil Khalidov, Marc Szafraniec, Seungeun Yi, Micha{\"e}l Ramamonjisoa, Francisco Massa, Daniel Haziza, Luca Wehrstedt, Jianyuan Wang, Timoth{\'e}e Darcet, Th{\'e}o Moutakanni, Leonel Sentana, Claire Roberts, Andrea Vedaldi, Jamie Tolan, John Brandt, Camille Couprie, Julien Mairal, Herv{\'e} J{\'e}gou, Patrick Labatut, and Piotr Bojanowski.
\newblock {{DINOv3}}, August 2025.

\bibitem[{\DJ}uki{\'c} et~al.(2025){\DJ}uki{\'c}, Lebailly, and Tuytelaars]{djukic2025object}
Nikola {\DJ}uki{\'c}, Tim Lebailly, and Tinne Tuytelaars.
\newblock Object-centric pretraining via target encoder bootstrapping.
\newblock In \emph{International Conference on Learning Representations}, volume 2025, pp.\  80781--80794, 2025.

\bibitem[Van~Assel et~al.(2026)Van~Assel, Ibrahim, Biancalani, Regev, and Balestriero]{van2026joint}
Hugues Van~Assel, Mark Ibrahim, Tommaso Biancalani, Aviv Regev, and Randall Balestriero.
\newblock Joint-embedding vs reconstruction: {{Provable}} benefits of latent space prediction for self-supervised learning.
\newblock \emph{Advances in neural information processing systems}, 38:\penalty0 21897--21937, 2026.

\bibitem[van~den Oord et~al.(2019)van~den Oord, Li, and Vinyals]{oord2019Representation}
Aaron van~den Oord, Yazhe Li, and Oriol Vinyals.
\newblock Representation {{Learning}} with {{Contrastive Predictive Coding}}, January 2019.

\bibitem[{Van der Walt} et~al.(2014){Van der Walt}, Sch{\"o}nberger, {Nunez-Iglesias}, Boulogne, Warner, Yager, Gouillart, and Yu]{van2014scikit}
Stefan {Van der Walt}, Johannes~L Sch{\"o}nberger, Juan {Nunez-Iglesias}, Fran{\c c}ois Boulogne, Joshua~D Warner, Neil Yager, Emmanuelle Gouillart, and Tony Yu.
\newblock Scikit-image: Image processing in {{Python}}.
\newblock \emph{PeerJ}, 2:\penalty0 e453, 2014.

\bibitem[Wang et~al.(2022)Wang, Wang, Wei, Yuille, and Shen]{wang2022cp}
Feng Wang, Huiyu Wang, Chen Wei, Alan Yuille, and Wei Shen.
\newblock Cp 2: {{Copy-paste}} contrastive pretraining for semantic segmentation.
\newblock In \emph{European Conference on Computer Vision}, pp.\  499--515. Springer, 2022.

\bibitem[Wang et~al.(2021)Wang, Zhang, Shen, Kong, and Li]{wang2021Dense}
Xinlong Wang, Rufeng Zhang, Chunhua Shen, Tao Kong, and Lei Li.
\newblock Dense contrastive learning for self-supervised visual pre-training.
\newblock In \emph{Proceedings of the {{IEEE}}/{{CVF}} Conference on Computer Vision and Pattern Recognition}, pp.\  3024--3033, 2021.

\bibitem[Wen et~al.(2022)Wen, Zhao, Zheng, Zhang, and Qi]{wen2022self}
Xin Wen, Bingchen Zhao, Anlin Zheng, Xiangyu Zhang, and Xiaojuan Qi.
\newblock Self-supervised visual representation learning with semantic grouping.
\newblock \emph{Advances in neural information processing systems}, 35:\penalty0 16423--16438, 2022.

\bibitem[Wen et~al.(2025)Wen, Zhao, Chen, Pang, and Qi]{wen2025DataCentric}
Xin Wen, Bingchen Zhao, Yilun Chen, Jiangmiao Pang, and Xiaojuan Qi.
\newblock A data-centric revisit of pre-trained vision models for robot learning.
\newblock In \emph{{{CVPR}}}, 2025.

\bibitem[Zhou et~al.(2017)Zhou, Zhao, Puig, Fidler, Barriuso, and Torralba]{zhou2017Scene}
Bolei Zhou, Hang Zhao, Xavier Puig, Sanja Fidler, Adela Barriuso, and Antonio Torralba.
\newblock Scene {{Parsing}} through {{ADE20K Dataset}}.
\newblock In \emph{2017 {{IEEE Conference}} on {{Computer Vision}} and {{Pattern Recognition}} ({{CVPR}})}, pp.\  5122--5130, July 2017.
\newblock \doi{10.1109/CVPR.2017.544}.

\bibitem[Zhou et~al.(2022)Zhou, Wei, Wang, Shen, Xie, Yuille, and Kong]{zhou2021IBOT}
Jinghao Zhou, Chen Wei, Huiyu Wang, Wei Shen, Cihang Xie, Alan Yuille, and Tao Kong.
\newblock {{iBOT}}: {{Image BERT}} pre-training with online tokenizer.
\newblock \emph{International Conference on Learning Representations (ICLR)}, 2022.

\end{thebibliography}
\bibliographystyle{iclr2027_conference}

\appendix
\newpage

\section{Implementation Details}
\subsection{Training Details}
\label{sec:training_details}
We list here the full set of hyperparameters used to train our main model. The encoder is a ViT-Base with a patch size of $16\times16$, trained from scratch with a stochastic-depth (drop-path) rate of $0.1$. The LeJEPA projection head is a three-layer MLP ($768\rightarrow2048\rightarrow2048\rightarrow64$) with synchronized batch normalization, projecting into a $64$-dimensional space, and the masked cross-attention pooling uses $8$ heads.

We train on COCO for $300$ epochs with AdamW at its default $\beta$ and $\epsilon$ with a learning rate of $5\cdot10^{-4}$, and a weight decay of $0.05$. The effective batch size is $256$ ($64$ images per GPU across $4$ NVIDIA RTX A6000 GPUs). The learning rate follows a cosine schedule decaying to $5\%$ of its peak, preceded by a single linear warm-up epoch. We use mixed-precision (AMP) training and clip gradients to a global norm of $1.0$. Per image we sample $2$ global views at $256\times256$ and $8$ local views at $128\times128$, together with the standard LeJEPA photometric augmentations (color jitter, grayscale, Gaussian blur, Gaussian noise, and solarization). More details on the copy-paste augmentation follow.

The training objective combines three terms: the LeJEPA SIGReg regularizer on the semantic object representation projections (weight $\lambda_{\mathrm{LeJEPA}}=0.05$), the LeJEPA-alignment loss (weight $1.0$), and the within-view instance-separation InfoNCE loss (weight $1.0$, temperature $0.1$). SIGReg uses $1024$ random projections and $17$ quadrature knots.

The SAM mask proposals are precomputed once with the SAM~2.1 Hiera-Large automatic mask generator \citep{ravi2025SAM}, run with $16$ points per side, a predicted-IoU threshold of $0.8$ (how confident SAM is in a mask), and a stability-score threshold of $0.92$ (how stable the binary mask is with respect to cutoff threshold variations). The other values are left to the default configuration. We cap the number of partitions (masks) per image at $64$, which bounds the per-batch object representation count and the cross-attention memory footprint.

The FH superpixel proposals are precomputed once with the Felzenszwalb--Huttenlocher graph-based segmentation \citep{felzenszwalb2004efficient}, in its scikit-image implementation \citep{van2014scikit}. It runs on each image at its original resolution in RGB. We set the granularity parameter to $k=255$ and leave the Gaussian pre-smoothing width ($\sigma=0.8$) and the minimum component size ($20$ pixels) at their defaults. The resulting superpixels over-segment with a median of $476$ segments per image. We cap the number of partitions per image at $64$ as we do for SAM. Since every image exceeds that, the cap does the actual selecting, and keeping the $64$ largest-area segments retains $83\%$ of the pixels on average (median $85\%$).

\subsection{Copy-paste Augmentations}
\label{sec:copy_paste_details}

Within each per-GPU batch of $64$ images we split the batch at random into $32$ hosts and $32$ donors and pair them one-to-one, so no image is both. For each pair we choose a single donor mask, sampled uniformly from those covering between $5\%$ and $50\%$ of the donor's first global view, a weak prior that excludes tiny, low-resolution objects and regions large enough to erase the host. The choice is made once per pair and reused for all $10$ views.

The copied region is transformed independently per view. It is mirrored horizontally with probability $0.5$, rotated uniformly in $[-25^\circ, 25^\circ]$ and scaled uniformly in $[0.7, 1.3]$, with the scale capped so that the warped region cannot exceed the $50\%$ area ceiling it was selected under. Pixels are resampled bilinearly and masks with nearest-neighbour interpolation, under identical parameters, so that image and mask cannot disagree. Regions of the frame opened up by the rotation are filled with the dataset mean colour, which attenuates the boundary artifacts a black fill would introduce without removing them entirely. The region is then translated by an integer offset drawn uniformly from the range that keeps its bounding box inside the frame. We redraw this offset up to $4$ times and accept the first placement whose intersection-over-union with the region's position in the donor's own frame is at most $0.1$, keeping the lowest-overlap draw of the four when no placement clears the threshold. Without this constraint the pasted patches retain the positional embeddings of the donor patches they are aligned against, and the pair differs only in context. If less than $1\%$ of the view remains covered after the transform, that view receives no paste.

Photometric augmentation is deferred until after compositing and applied once per view, jointly over the host and the pasted region. Applying it before the paste leaves each side carrying its own colour and blur draw, which makes the paste boundary detectable from the mismatch alone and leaves the host-side pasted pixels identical to the donor pixels they are aligned against.

\subsection{Downstream Task Details}
\label{sec:downstream_details}
All encoders are frozen and used purely as feature extractors. For each dataset we run a single forward pass per image and cache its outputs, so every probe on a given dataset reads the same features. We work with three kinds of representations. \emph{Patch features} are the last-layer feature maps. At a patch size of $16\times16$ a $512\times512$ image yields a $32\times32$ grid. The \emph{image-level} representation is the mean of all patch features over the spatial grid. For the image-level models, Image LeJEPA and DINOv3, we additionally extract the \texttt{[cls]} token. The \emph{object-level} representation of a mask is the weighted mean of the patch features over the patches. The weights for this mean come from the patch-wise average-pooled mask. For our method we additionally evaluate the semantic object representations $\mathbf{z}_{v,k}$. Each dataset is processed at a fixed input resolution, always a multiple of the $16$-pixel patch size: ADE20k, COCO, and NAVI at $512\times512$ and ImageNet-1k at $224\times224$. DAVIS instead keeps its native $480$p aspect ratio, resizing each frame so that its shorter side is $512$ pixels (rounded to a multiple of $16$).

\paragraph{Instance Clustering (ADE20k)}
For each validation image we $\ell_2$-normalize the foreground patch features and run K-Means with $K$ set to the number of ground-truth foreground instances in that image. We use $5$ random initializations per image. The adjusted rand index (FG-ARI) is computed directly over all foreground patches. Metrics are averaged over all validation images containing at least two instances. We use the 2021 ADE20k release, which provides true per-instance masks.

\paragraph{Same-Instance Quadratic Probe (ADE20k)}
Following \citet{li2026Does}, whose notation we adopt in this paragraph independently of the rest of the paper, we learn
\begin{align}
    \mathrm{IsSameObject}(x,y)=\sigma\!\left((Px)^{\top}\mathrm{diag}(g)(Py)+b\right)
\end{align}
that predicts, for a pair of raw (unnormalized) patch embeddings, whether they belong to the same object instance. Here $P\in\mathbb{R}^{k\times d}$ projects each patch into a $k=32$ dimensional binding subspace and $g\in\mathbb{R}^{k}$ is a learned signed signature, giving a symmetric, rank-$\leq k$ form with $O(kd)$ parameters that is not constrained to be positive semi-definite. The probe is trained on the ADE20k training split with binary cross-entropy over all within-image off-diagonal patch pairs, sampling at most $64$ foreground patches per image. We optimize with Adam (learning rate $10^{-3}$, weight decay $10^{-4}$, cosine schedule) for $5$ epochs at $32$ images per batch. In the evaluation stage, we score every foreground patch pair (up to $256$ patches per validation image). We calculate the threshold-free ROC-AUC on all patch pairs that belong to the same category for each image and then average over all validation images.

\paragraph{Semantic Segmentation (ADE20k)}
We train a single $1\times1$ convolutional head on the frozen patch features to predict the $150$ ADE20k semantic classes, with cross-entropy and the background class ignored. Training is done at the feature-grid resolution (the label mask is nearest-neighbour downsampled), and at evaluation the logits are bilinearly upsampled back to the cached mask resolution before the argmax, so the metric follows the full-resolution protocol. We optimize with Adam (learning rate $10^{-2}$, cosine schedule) for $5$ epochs at a batch size of $128$. We report mean IoU (over the classes present in each image) averaged over the validation images.

\paragraph{Label Propagation (DAVIS)}
We follow the label-propagation tracker of \citet{caron2021Emerging}. The ground-truth mask of the first frame is propagated to subsequent frames by affinity-weighted voting over $\ell_2$-normalized patch features: affinities are softmax-weighted with temperature $0.1$, restricted to a spatial neighborhood of radius $12$ patches around each query, and sparsified to the top $5$ source patches per query. The context set for each frame is the first frame plus the $7$ most recently predicted frames. Frames keep their aspect ratio (short-side resize) and are processed one at a time. We report $\mathcal{J}\&\mathcal{F}$, the mean of region similarity $\mathcal{J}$ (mask IoU) and contour accuracy $\mathcal{F}$ (boundary F-measure, tolerance $0.8\%$ of the image diagonal), averaged first over objects and frames within a video, then over videos.

\paragraph{Object Classification (ADE20k)}
Each object is represented by its object-level descriptor, extracted with its ground-truth mask, and for our method additionally by its semantic object representation. We train a linear head on the training-split objects with cross-entropy, using Adam (learning rate $10^{-2}$, cosine schedule) for $50$ epochs at a batch size of $4096$. The classes present in the training split are densely re-indexed, and validation objects of classes unseen in training are excluded from scoring. We report top-1 balanced accuracy over the validation objects.

\paragraph{Object Re-Identification (NAVI)}
On the NAVI wild set, each object instance appears in multiple images under varying background, pose, and lighting. We represent each object by its mask-pooled patch descriptor (and, for our method, its semantic object representation), $\ell_2$-normalized. We build a memory bank by sampling $5$ representations per identity and classify every remaining image by its cosine-nearest exemplar across all identities. Identities with $\leq 5$ images are skipped. We repeat the sampling $10$ times and report the mean balanced accuracy.

\paragraph{Image Classification (ImageNet-1k)}
The object-centric models (SlotMIM and our method) do not learn a usable \texttt{[cls]} token, so we represent each of their images by the image-level descriptor (the global average of its patch features). For the image-level models, such as Image LeJEPA and DINOv3, we additionally report results using their \texttt{[cls]} token. On the $\ell_2$-normalized, standardized representations we fit a multinomial logistic-regression probe (inverse regularization $C=1$) on the full training split and report top-1 balanced accuracy on the validation split.

\section{Further Results}
\subsection{Results on Smaller Backbones}
\label{sec:smaller_archs}

\begin{table}[h]
\centering
\footnotesize
\setlength{\tabcolsep}{3pt}
\caption{Results with frozen encoders on smaller backbones, extending Table~\ref{tab:main_results} to methods whose authors release checkpoints only for ViT-S/16 or ResNet-50. Blocks group methods pre-trained on COCO with ground-truth instance masks on ViT-S/16 (top), without annotations on ViT-S/16 (middle), and without annotations on ResNet-50 (bottom). Bold marks the best value per block. CrIBo, OCEBO, DenseCL and SlotCon are evaluated with the authors' checkpoints. DetCon, Image LeJEPA and our method are trained by us for 150 epochs. Readouts as in Table~\ref{tab:main_results}: $^{a}$~semantic object representation $\mathbf{z}$, $^{b}$~\texttt{[cls]} token. ResNet-50 features have stride 32 rather than 16, so their dense results are not directly comparable to the ViT rows.}
\label{tab:more_main_results}
\resizebox{\textwidth}{!}{%
\begin{tabular}{@{}lrlll@{\hskip 10pt}ccccccc@{}}
\toprule
\shortstack{{}\\{}\\\textbf{Method}} & \shortstack{{}\\{}\\\textbf{Ep}} & \shortstack{{}\\{}\\\textbf{Architecture}} & \shortstack{{}\\{}\\\textbf{Dataset}} & \shortstack{{}\\\textbf{GT}\\\textbf{source}} & \shortstack{\textbf{ADE20k}\\\textbf{Instance Clust.}\\\textbf{FG-ARI}} & \shortstack{\textbf{ADE20k}\\\textbf{Same-Class Pair}\\\textbf{AUC}} & \shortstack{\textbf{ADE20k}\\\textbf{Linear Seg.}\\\textbf{mIoU}} & \shortstack{\textbf{DAVIS}\\\textbf{Tracking}\\$\bm{\mathcal{J}\&\mathcal{F}}$} & \shortstack{\textbf{ADE20k}\\\textbf{Object Cls.}\\\textbf{Top-1 Bal. Acc.}} & \shortstack{\textbf{NAVI}\\\textbf{Re-ID}\\\textbf{5-shot Bal. Acc.}} & \shortstack{\textbf{ImageNet-1k}\\\textbf{Image Cls.}\\\textbf{Top-1 Bal. Acc.}} \\
\midrule
DetCon & 150 & ViT-S/16 & COCO & instances & 0.304 & 0.805 & 0.348 & 0.613 & 0.451 & 0.572 & 0.451 \\
Ours & 150 & ViT-S/16 & COCO & instances & \textbf{0.377} & \textbf{0.847} & \textbf{0.385} & \textbf{0.676} & \textbf{0.525}$^{a}$ & \textbf{0.778}$^{a}$ & \textbf{0.512} \\
\midrule
CrIBo & 300 & ViT-S/16 & COCO & / & \textbf{0.324} & \textbf{0.814} & \textbf{0.373} & \textbf{0.617} & \textbf{0.453} & 0.539 & \textbf{0.479} \\
Image LeJEPA & 150 & ViT-S/16 & COCO & / & 0.283 & 0.786 & 0.319 & 0.608 & 0.387 & \textbf{0.561} & 0.436$^{b}$ \\
OCEBO & 400 & ViT-S/16 & COCO & / & 0.323 & 0.806 & 0.335 & 0.591 & 0.372 & 0.449 & 0.417 \\
\midrule
DenseCL & 800 & ResNet-50 & COCO & / & 0.288 & 0.769 & 0.302 & \textbf{0.357} & 0.385 & 0.532 & 0.427 \\
SlotCon & 800 & ResNet-50 & COCO & / & \textbf{0.299} & \textbf{0.772} & \textbf{0.339} & 0.315 & \textbf{0.424} & \textbf{0.635} & \textbf{0.468} \\
\bottomrule
\end{tabular}
}
\par
\end{table}

Table~\ref{tab:more_main_results} extends the comparison of Table~\ref{tab:main_results} to methods whose authors release checkpoints only for smaller backbones. On ViT-S/16 we compare against CrIBo~\citep{lebailly2024cribo} and OCEBO~\citep{djukic2025object}, using the authors' COCO checkpoints, and against DetCon and image-level LeJEPA trained by us with the same recipe as in the main text. Our ViT-S model, trained for 150 epochs, leads every column of the ViT-S block, including against CrIBo at 300 and OCEBO at 400 epochs, with the largest margins on the instance-level tasks (FG-ARI $0.377$ vs.\ $0.324$, NAVI $0.778$ vs.\ $0.539$). The ResNet-50 rows, DenseCL~\citep{wang2021Dense} and SlotCon~\citep{wen2022self}, are listed for completeness. Their stride-32 feature maps are coarser than the $16\times16$ patch grid of the ViTs, so their dense results are not directly comparable and we draw no conclusion from them.

\subsection{Epoch Ablation}
Table~\ref{tab:epoch_ablation} compares our model at 150 and 300 epochs. All metrics change by at most $0.008$, so the 150-epoch runs of Table~\ref{tab:ablation} and Figure~\ref{fig:dataset-size-ablation} are representative of the main model.
\begin{table}[h]
\centering
\footnotesize
\setlength{\tabcolsep}{3pt}
\caption{Epoch ablation for our method on full COCO at 150 and 300 epochs.}
\label{tab:epoch_ablation}
\resizebox{\textwidth}{!}{%
\begin{tabular}{@{}l@{\hskip 10pt}ccccccc@{}}
\toprule
\shortstack{{}\\{}\\\textbf{Budget}} & \shortstack{\textbf{ADE20k}\\\textbf{Instance Clust.}\\\textbf{FG-ARI}} & \shortstack{\textbf{ADE20k}\\\textbf{Same-Class Pair}\\\textbf{AUC}} & \shortstack{\textbf{ADE20k}\\\textbf{Linear Seg.}\\\textbf{mIoU}} & \shortstack{\textbf{DAVIS}\\\textbf{Tracking}\\$\bm{\mathcal{J}\&\mathcal{F}}$} & \shortstack{\textbf{ADE20k}\\\textbf{Object Cls.}\\\textbf{Top-1 Bal. Acc.}} & \shortstack{\textbf{NAVI}\\\textbf{Re-ID}\\\textbf{5-shot Bal. Acc.}} & \shortstack{\textbf{ImageNet-1k}\\\textbf{Image Cls.}\\\textbf{Top-1 Bal. Acc.}} \\
\midrule
150 ep & 0.422 & \textbf{0.851} & 0.401 & \textbf{0.691} & \textbf{0.563} & \textbf{0.783} & 0.572 \\
300 ep & \textbf{0.423} & 0.850 & \textbf{0.403} & 0.683 & 0.555 & 0.775 & \textbf{0.574} \\
\midrule
$\Delta$ & $+0.001$ & $-0.001$ & $+0.002$ & $-0.008$ & $-0.008$ & $-0.008$ & $+0.002$ \\
\bottomrule
\end{tabular}
}
\end{table}

\subsection{Loss Regimes across Dataset Sizes}
\label{sec:regime_transition}
Figure~\ref{fig:dataset-size-ablation} shows a non-monotonic NAVI curve for our method: re-identification is high on the smallest subsets, dips at $10\%$ of COCO, and recovers at full scale ($0.737 \to 0.651 \to 0.783$ for ViT-B). The semantic tasks show no such dip. They improve steadily within each compute regime and jump at the transition (ImageNet $0.373 \to 0.483$, ADE20k $0.325 \to 0.368$ from $1024$ images at $15$k steps to $11.8$k images at $69$k steps). We relate this to which of the two losses shapes the representation at each scale.

On $256$ images the alignment term of $\mathcal{L}_{\mathrm{ObjectLeJEPA}}$ converges to $0.006$ (SIGReg: $0.001$), an order of magnitude below the $0.075$ ($0.003$) reached on full COCO: with few images and objects, the encoder can satisfy cross-view agreement almost exactly, and once the alignment term is near zero its gradient, which scales with the residual, vanishes. The instance loss, which must separate every patch from all other patches in its view, remains far from its minimum at both scales ($2.62$ on $256$ images, $2.88$ on full COCO). On tiny datasets the training signal therefore comes almost entirely from instance separation, which favours tracking and re-identification and explains why our method reaches its strongest relative advantage on DAVIS and NAVI with $256$--$1024$ images. As the dataset grows, the alignment term can no longer be solved by memorization and starts shaping the representation as well. The representation turns more semantic, which lifts ImageNet and ADE20k and, at $10\%$ of COCO, costs re-identification. At full scale the encoder captures both, and the representation is semantic and instance-specific at once (NAVI $0.783$).

\section{Qualitative Results}
\subsection{SlotMIM Partitions}
\begin{figure*}[h]
  \centering
  \setlength{\tabcolsep}{0pt}
  \begin{tabular}{c}
    \includegraphics[width=\textwidth, trim=10 5 10 5, clip]{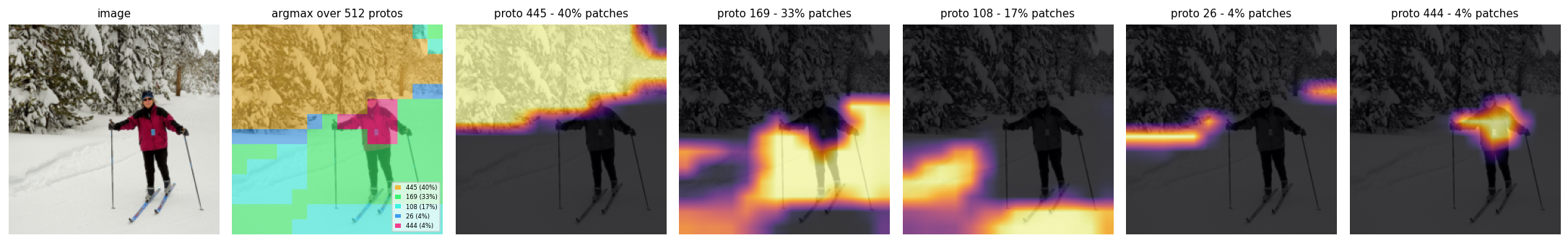} \\[-2pt]
    \includegraphics[width=\textwidth, trim=10 5 10 5, clip]{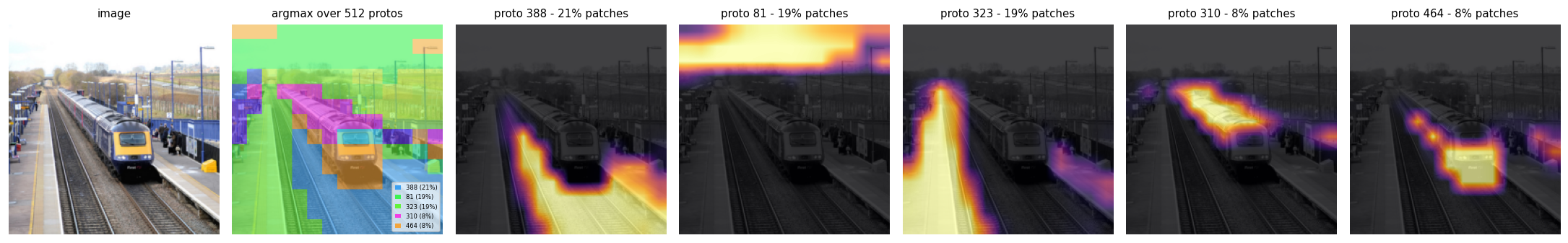} \\[-2pt]
    \includegraphics[width=\textwidth, trim=10 5 10 5, clip]{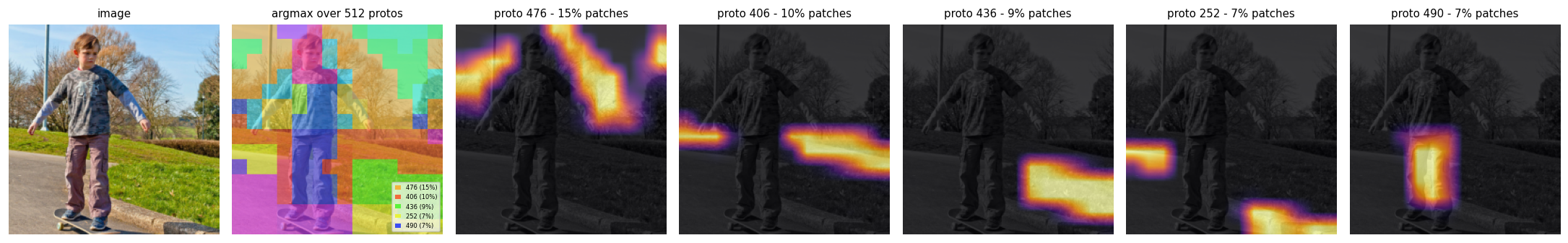} \\[-2pt]
    \includegraphics[width=\textwidth, trim=10 5 10 5, clip]{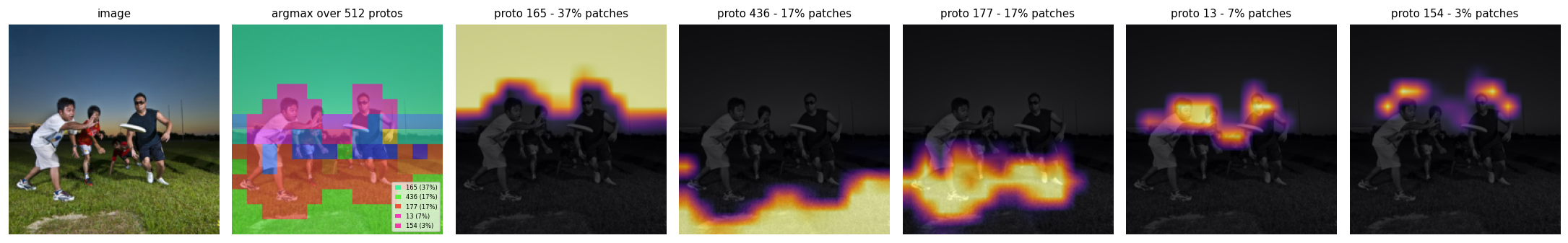} \\[-2pt]
    \includegraphics[width=\textwidth, trim=10 5 10 5, clip]{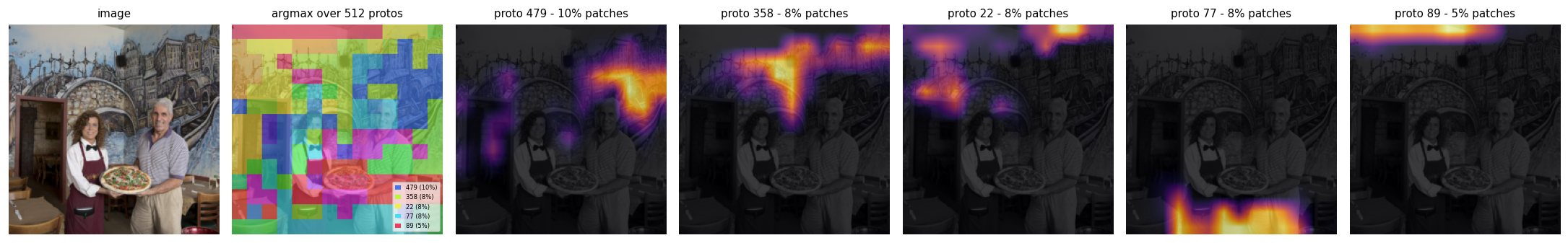}
  \end{tabular}
  \caption{Partitions learned by SlotMIM (ViT-B/16, 512 prototypes, 800 epochs on COCO) on five COCO \texttt{val2017} images. The first column depicts the input at a resolution of $224\times224$. The second column visualizes the patches colored by the prototype with the highest cosine similarity. The five most frequent prototypes and their share of patches are in the legend. The remaining columns show the soft assignment of patches to those five prototypes (softmax over prototypes at temperature $0.07$, each panel normalized to its own maximum). Prototype colors are consistent across rows.}
  \label{fig:slotmim-partitions}
\end{figure*}
In our work, we want to investigate how we can leverage object compositionality to train image encoders with fewer images. We argue that many object-centric models that try to learn partitioning and representation at the same time, cannot be used to investigate this since their partitioning does not align with object decomposition. SlotMIM assigns each patch to a global prototype. If there are two instances of the same category in an image, they should arguably be assigned to the same prototype, which leads to two instances sharing one local object representation defeating the premise of object-centric representations. For each image we take the patch tokens of the EMA teacher, pass them through its projector and compute cosine similarities to the 512 teacher prototypes. The hard partition assigns every patch to its highest-scoring prototype. The soft maps are the softmax over prototypes at the training temperature of $0.07$, shown for the five prototypes that claim the most patches. Both are computed at the training resolution and upsampled for display. Figure~\ref{fig:slotmim-partitions} illustrates the failure mode: the resulting partitions do not separate instances of the same category. This matches SlotMIM's weak NAVI re-identification and instance-clustering results and motivates assuming masks that align with objects.

\clearpage
\subsection{Visual Results}
In this section we present visual results for models that are trained on the COCO dataset. Only DINOv3 was trained on a larger dataset. Since COCO does not include images of fish, it is interesting to investigate the encoders' instance-separating and cross-image matching capabilities on out-of-distribution objects. We select a patch feature belonging to a fish in one image and visualize the feature similarity (cosine-similarity) to other patches from the same image and other images from ImageNet. Figure~\ref{fig:anchor-qual} shows how our method demonstrates much higher instance separation than other methods within image 1. Our method clearly separates the fish in the foreground from the background even though the model has never seen fish during training. R-MAE seems to rely on low-level features since it does not provide a clean similarity map when the fish has a different color (yellow fish in image 2) and it assigns high similarities to the flowers that are similar color-wise to the anchor fish. This is the case with all COCO-trained models and only DINOv3 assigns them low similarity.

\begin{figure*}[t]
  \centering
  \resizebox{\textwidth}{!}{%
  \begin{tikzpicture}[font=\small]
    \def\cw{3}\def\g{0.08}
    \pgfmathsetmacro\step{\cw+\g}
    \foreach \r/\m/\name in {%
        0/original/{input},
        1/imagelejepa/{Image\\LeJEPA},
        2/slotmim800/{SlotMIM},
        3/rmae/{R-MAE},
        4/detcon/{DetCon},
        5/ours/{Ours},
        6/dinov3/{DINOv3}}{
      \node[rotate=90,align=center] at (-1.95,-\r*\step) {\name};
      \foreach \c/\i in {0/53,1/56,2/63,3/74,4/96}{
        \node[inner sep=0] at (\c*\step,-\r*\step)
          {\includegraphics[width=\cw cm]{resources/output/anchor_53_x14_y14/\m_index\i.png}};
      }
    }
    \foreach \c/\h in {0/{image 1 (anchor)},1/{image 2},2/{image 3},3/{image 4},4/{image 5}}{
      \node[anchor=south] at (\c*\step,0.5*\cw+0.15) {\h};
    }
  \end{tikzpicture}}
  \caption{Cosine similarity of all patches to one anchor patch (white ring, image~1), across five goldfish images (columns) and backbones (rows). Similarities are shown in grayscale, black $=0$ to white $=1$.}
  \label{fig:anchor-qual}
\end{figure*}

\clearpage
We also provide per-image PCA visualizations of the patch features on 5 images from the COCO validation set in Figure~\ref{fig:pca-qual}. Different instances are cleanly separable (3 zebras in the first column, buses in the second column, car in the background in the third column), while Section~\ref{sec:downstream_tasks} shows that the patch features are also semantically meaningful.

\begin{figure*}[t]
  \centering
  \resizebox{\textwidth}{!}{%
  \begin{tikzpicture}[font=\small]
    \def\cw{3}\def\g{0.08}
    \pgfmathsetmacro\step{\cw+\g}
    \foreach \r/\m/\name in {%
        0/original/{input},
        1/imagelejepa/{Image\\LeJEPA},
        2/slotmim800/{SlotMIM},
        3/rmae/{R-MAE},
        4/detcon/{DetCon},
        5/ours/{Ours},
        6/dinov3/{DINOv3}}{
      \node[rotate=90,align=center] at (-1.95,-\r*\step) {\name};
      \foreach \c/\i in {0/000000011760,1/000000090208,2/000000293071,
                         3/000000312192,4/000000521231}{
        \node[inner sep=0] at (\c*\step,-\r*\step)
          {\includegraphics[width=\cw cm]{resources/output/pca_fig/\m_\i.png}};
      }
    }
  \end{tikzpicture}}
  \caption{First three principal components of the frozen patch features, mapped to RGB, on five COCO \texttt{val2017} images (columns) across backbones (rows). The basis is fitted independently on that image's $\ell_2$-normalized patch features, so colors are arbitrary and not comparable across cells. What is comparable is how sharply each backbone's leading components align with object extent.}
  \label{fig:pca-qual}
\end{figure*}

\clearpage
\end{document}